\documentclass{article} 
\usepackage{iclr2027_conference,times}

\usepackage{amsmath,amsfonts,bm}

\def\eqref#1{equation~\ref{#1}}

\def\1{\bm{1}}

\DeclareMathAlphabet{\mathsfit}{\encodingdefault}{\sfdefault}{m}{sl}
\SetMathAlphabet{\mathsfit}{bold}{\encodingdefault}{\sfdefault}{bx}{n}

\iclrfinalcopy 

\usepackage{xcolor,color,soul}
\usepackage{tabularx}
\colorlet{soulblue}{blue!20}
\newif\iffinal
\finaltrue 
\iffinal
  \newcommand{\yw}[1]{}
  \newcommand{\ap}[1]{}
  \newcommand{\lc}[1]{}
\else
  \newcommand{\yw}[1]{\sethlcolor{soulblue}\hl{[Yvonne: #1]}}
  \newcommand{\ap}[1]{\sethlcolor{yellow}\hl{[Arvind: #1]}}
  \newcommand{\lc}[1]{\sethlcolor{pink}\hl{[Leo: #1]}}
\fi

\newcommand{\systemname}{SLIP}
\usepackage[utf8]{inputenc} 
\usepackage[T1]{fontenc}    
\usepackage{hyperref}       
\hypersetup{colorlinks=true, allcolors=[rgb]{0,0.08,0.45}} 
\usepackage{url}            
\usepackage{booktabs}       
\usepackage{placeins}
\usepackage{amsfonts}       
\usepackage{amsmath}        
\usepackage{amssymb}        
\usepackage{nicefrac}       
\usepackage{microtype}      
\usepackage{graphicx}

\usepackage{multirow}
\usepackage{array}
\usepackage[table]{xcolor} 
\usepackage{pifont}
\newcommand{\cmark}{\textcolor{green!60!black}{\ding{51}}}
\newcommand{\xmark}{\textcolor{red!70!black}{\ding{55}}}
\usepackage{float}
\usepackage{fontawesome5}
\usepackage{algorithm}
\usepackage{enumitem}
\usepackage{subcaption}   
\usepackage{tabularx}  
\usepackage{longtable}
\usepackage{siunitx}   
\usepackage{caption}   
\usepackage{listings}
\usepackage{upquote}  
\usepackage{xcolor,colortbl}
\usepackage{tablefootnote}
\usepackage{adjustbox}

\newcommand{\srcone}[1]{\vtop{\hbox{#1}\hbox{}\hbox{}}}
\newcommand{\srcthree}[3]{\vtop{\hbox{#1}\hbox{#2}\hbox{#3}}}
\newcolumntype{L}[1]{>{\raggedright\arraybackslash}p{#1}}
\lstdefinestyle{overleaf}{
    backgroundcolor=\color[rgb]{0.95,0.95,0.92},   
    commentstyle=\color[rgb]{0,0.6,0},
    keywordstyle=\color{magenta},
    numberstyle=\tiny\color[rgb]{0.5,0.5,0.5},
    stringstyle=\color[rgb]{0.58,0,0.82},
    basicstyle=\ttfamily\footnotesize,
    breakatwhitespace=false,         
    breaklines=true,                 
    captionpos=b,                    
    keepspaces=true,                 
    numbers=left,                    
    numbersep=5pt,                  
    showspaces=false,                
    showstringspaces=false,
    showtabs=false,                  
    tabsize=2
}
\lstdefinestyle{mocov3}{
  backgroundcolor=\color{white},
  basicstyle=\fontsize{7.5pt}{7.5pt}\ttfamily\selectfont,
  columns=fullflexible,
  breaklines=true,
  captionpos=b,
  commentstyle=\fontsize{7.5pt}{7.5pt}\color[rgb]{0.25,0.5,0.5},
  keywordstyle=\fontsize{7.5pt}{7.5pt}\color[rgb]{0.85,0.18,0.50},
  numbers=none,
}
\usepackage{etoolbox}
\makeatletter
\AfterEndEnvironment{algorithm}{\let\@algcomment\relax}
\AtEndEnvironment{algorithm}{\kern2pt\hrule\relax\vskip3pt\@algcomment}
\let\@algcomment\relax
\newcommand\algcomment[1]{\def\@algcomment{\footnotesize#1}}
\renewcommand\fs@ruled{\def\@fs@cfont{\bfseries}\let\@fs@capt\floatc@ruled
  \def\@fs@pre{\hrule height.8pt depth0pt \kern2pt}%
  \def\@fs@post{}%
  \def\@fs@mid{\kern2pt\hrule\kern2pt}%
  \let\@fs@iftopcapt\iftrue}
\makeatother
\definecolor{headerblue}{HTML}{EBF3FF}
\definecolor{textred}{HTML}{980000}
\definecolor{textgreen}{HTML}{274E13}
\definecolor{mygray}{gray}{0.5}
\newcommand{\negval}[1]{\textcolor{textred}{\scriptsize (#1)}}
\newcommand{\posval}[1]{\textcolor{textgreen}{\scriptsize (#1)}}

\title{Learning Transferable Sensor Models \\[0.15em] via Language-Informed Pretraining}

\author{%
  Yuliang Chen, Arvind Pillai, Yu Yvonne Wu, Sudarshan Regmi, Tess Z. Griffin, \\
  \textbf{Lisa Marsch, Michael V. Heinz, Nicholas C. Jacobson, Andrew Campbell} \\
  Dartmouth College \\
  \texttt{yuliang.chen.gr@dartmouth.edu} \\[6pt]
  \normalfont
  \raisebox{-0.2\height}{\includegraphics[height=1.1em]{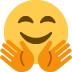}}\,\textbf{\textsc{Model}}: \href{https://huggingface.co/LeoChen085/SLIP}{\texttt{\textcolor{blue}{LeoChen085/SLIP}}}
  \qquad
  \faGithub\,\textbf{\textsc{Code}}: \href{https://github.com/yuc0805/SLIP}{\texttt{\textcolor{blue}{yuc0805/SLIP}}}
}

\begin{document}

\maketitle
\lhead{Preprint}


\begin{abstract}

Multimodal language models have demonstrated strong semantic understanding and reasoning over physiological and behavioral signals captured from diverse healthcare sensors. However, existing sensor-language models can only process a fixed number of sensor channels at a fixed sampling rate, making them difficult to adapt to new or different sensor configurations. This inflexibility prevents models from generalizing across heterogeneous health sensing ecosystems, where data spans high-frequency wearable streams to sparse, day-scale clinical signals. To bridge this gap, we introduce \textbf{SLIP} (\textbf{S}ensor \textbf{L}anguage-\textbf{I}nformed \textbf{P}retraining), an open-source framework for learning language-aligned representations that generalize across diverse sensor setups. SLIP integrates contrastive alignment with sensor-conditioned captioning, facilitating both discriminative understanding and generative reasoning. By repurposing a pretrained decoder-only language model via cross-attention and introducing a flexible patch embedder, \systemname{} transfers to new sensor configurations at inference without retraining, regardless of their native sampling rate or input length. Across 11 datasets, SLIP demonstrates superior performance in retrieval, signal captioning, and question answering. It achieves a 77.14\% average linear-probing accuracy, a 3.56\% improvement over the strongest baseline (SensorLM). Beyond classification, \systemname{} supports open-vocabulary sensor captioning and question answering without any architectural modifications. Our work offers broad implications for the development of sensor-language models, utilizing cross-modal supervision to ensure more robust generalizability in health sensing applications.
\end{abstract}

\section{Introduction}\label{introduction}

Healthcare relies on a diverse array of sensors, from wearables to clinical monitors, which capture longitudinal physiological and behavioral data critical for automated diagnosis and intervention. To enable scalable analysis of these signals, self-supervised learning (SSL) has become the dominant pretraining paradigm for learning general-purpose representations from unlabeled sensor data. 
Recent work shows that such pretraining improves a range of health applications spanning activity and sleep monitoring, stress recognition, and clinical diagnosis~\citep{spathis2022longitudinal, thapa2026multimodal, Li_2025}. 
However, because sensor modalities differ fundamentally in sampling rate, channel structure, and signal dynamics, many such efforts are restricted to a single modality such as ECG or PPG~\citep{pillai2024papagei, saha2025pulse, mckeen2025ecg}, achieving strong in-domain performance but failing to generalize across the heterogeneous sensor configurations encountered in real-world deployments.
NormWear~\citep{luo2025normwear} handles an arbitrary number of channels and sensor modalities, but requires a fixed sampling rate and input length.
In parallel, general-purpose time series foundation models~\citep{chronos2,goswami2024moment} pretrain on mixture corpora spanning multiple sensor modalities with predictive SSL objectives. However, these models are optimized for forecasting, and whether their learned representations transfer to downstream classification in domains such as human activity recognition, clinical diagnosis, and physiological monitoring remains largely unexplored (Figure~\ref{fig:chronos2_demo}).

Recently, large language models (LLMs) have showcased strong semantic abstraction and generalization in zero-shot and few-shot settings. Consequently, recent work explores aligning LLMs with time series and sensor representations to overcome the limitations of reconstruction-based or sensor-specific pretraining. For instance, HealthLLM~\citep{kim2024healthllmlargelanguagemodels} reformulates sensor signals as text to leverage pretrained LLMs; however, this often leads to substantial semantic and temporal information loss. To better preserve signal structure, recent methods map time series into the LLM embedding space via projection layers and rely on the language model for contextual inference~\citep{jin2023time,chatts,itformer}. A complementary line of work uses contrastive cross-modal learning frameworks to align sensor and text modalities, enabling zero-shot retrieval and captioning~\citep{ndir2025eegcliplearningeeg, sensorlm}. Despite offering a promising pretraining paradigm, most of these methods fix the sensor encoder to a specific channel layout and sampling rate, so the pretrained model cannot be applied to data with a different input configuration without retraining~\citep{sensorlm, itformer}. More fundamentally, contrastive learning benefits from broad data diversity~\citep{radford2021clip}, yet these models pretrain on single-domain corpora, constraining the representation space they can learn.

To address these limitations, we propose \systemname{}, a unified sensor-language encoder that learns transferable representations via contrastive alignment across heterogeneous sensor configurations. \systemname{} departs from prior sensor-language models in three ways.
First, we pretrain on a mixture corpus spanning multiple sensor domains, making the model sensor-agnostic: it assumes no fixed channel set or device type. Cross-domain negatives force the contrastive objective to align on semantic content rather than domain-specific signal characteristics.
Second, this diversity means samples vary widely in sampling rate and length, and high-frequency modalities such as ECG at 500\,Hz impose significant compute under a fixed patch size. To address this, we design \textit{FlexMLP}, a patch embedder parameterized at a base patch size and resized on the fly per sample, making \systemname{} \emph{length-adaptive}: a single checkpoint adjusts its patch size to each input's sampling rate and length, without resampling or retraining.
Third, rather than training from scratch, we initialize \systemname{} from a pretrained decoder-only LM, partitioning its layers into a text encoder and a generative decoder. The resulting encoder-decoder supports both sensor-language understanding and open-vocabulary generation, unifying capabilities that prior work addresses separately. We summarize these differences in Table~\ref{tab:capability_compare} and list our contributions below:
\begin{figure}[t]
\centering
\begin{minipage}[t]{0.54\linewidth}
    \centering
    \captionof{figure}{\textbf{Illustrated example of the forecasting-classification gap}. Chronos-2 achieves low reconstruction error on UCI-HAR (MSE = 0.96), yet its latent representations fail to discriminate between semantically distinct activities (walking downstairs vs.\ upstairs). This shows that predictive SSL does not necessarily yield representations that transfer to downstream classification tasks.}
    \includegraphics[width=\linewidth]{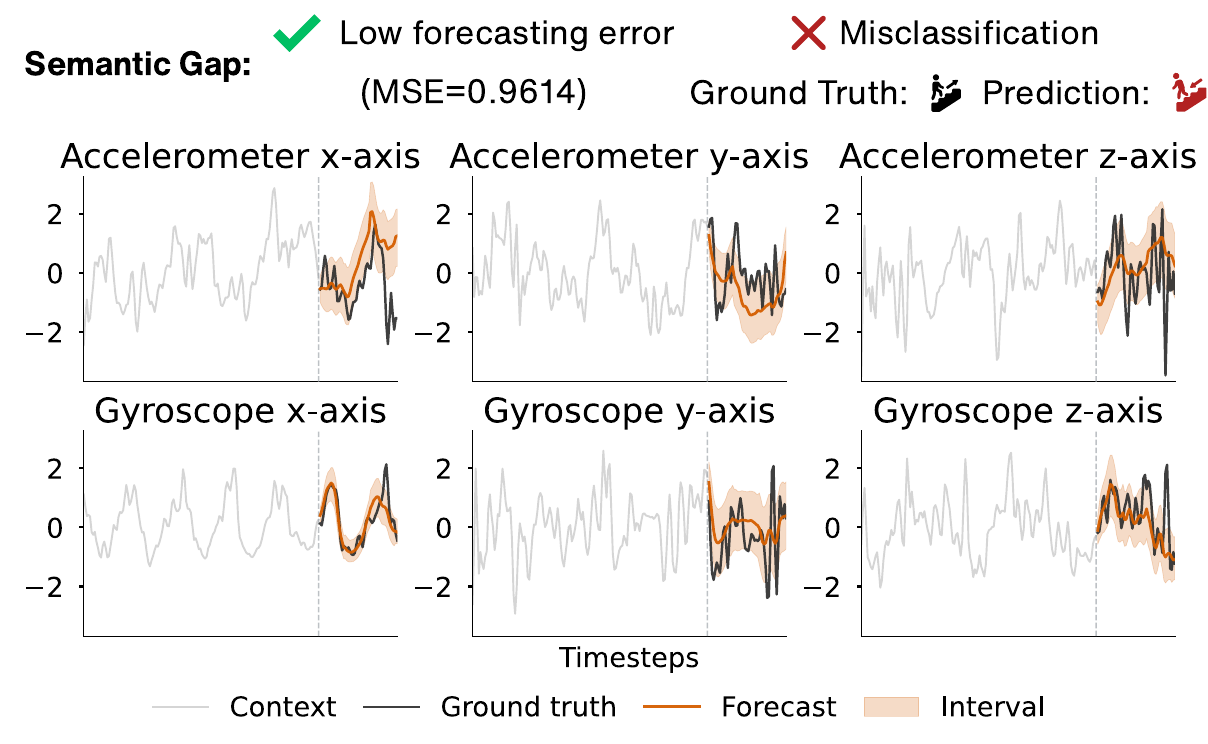}
    \label{fig:chronos2_demo}
\end{minipage}%
\hfill
\begin{minipage}[t]{0.42\linewidth}
    \centering
    \footnotesize
    \captionof{table}{\textbf{Comparison of sensor-text modeling approaches.} We compare prior work and SLIP across key capabilities, including \textbf{length-adaptive} encoding (patch size is adjusted at inference to match the input length and sampling rate, requiring no retraining), \textbf{sensor-agnostic} input (no fixed sensor set required), cross-modal \textbf{retrieval}, and sensor \textbf{question answering (QA)}.}
    \vspace{-1pt}
    \label{tab:capability_compare}
    \renewcommand{\arraystretch}{1.25}
    \setlength{\tabcolsep}{2pt}
    \resizebox{\linewidth}{!}{%
    \begin{tabular}{l c c c c}
    \toprule
    \textbf{Study} &
    \textbf{\shortstack{Length \\ Adaptive}} &
    \textbf{\shortstack{Sensor \\ Agnostic}} &
    \textbf{Retrieval} &
    \textbf{QA} \\
    \midrule
    Chronos \citep{ansari2024chronos}    & \xmark & \cmark & \xmark & \xmark \\
    Chronos2  \citep{chronos2}           & \xmark & \cmark & \xmark & \xmark \\
    Sundial  \citep{liu2025sundial}      & \xmark & \cmark & \xmark & \xmark \\
    SensorLM  \citep{sensorlm}           & \xmark & \xmark & \cmark & \xmark \\
    NormWear  \citep{luo2025normwear}    & \xmark & \cmark & \cmark & \xmark \\
    ChatTS  \citep{chatts}             & \xmark & \cmark & \xmark & \cmark \\
    ITFormer  \citep{itformer}           & \xmark & \xmark & \xmark & \cmark \\
    OpenTSLM   \citep{langer2025opentslm} & \xmark & \xmark & \xmark & \cmark \\
    \midrule
    \rowcolor{headerblue}
    \textbf{SLIP (ours)} & \cmark & \cmark & \cmark & \cmark \\
    \bottomrule
    \end{tabular}%
    
    }
\end{minipage}
\vspace{-0.6cm}
\end{figure}
\begin{itemize}
    \item \textbf{Unified and transferable sensor-language model:} We show that contrastive pretraining on a multi-domain mixture corpus yields sensor representations that transfer across heterogeneous configurations. With \textit{FlexMLP} to handle varying sampling rates and lengths, a single checkpoint supports classification, retrieval, captioning, and QA across sensor domains without resampling or retraining.
    \item \textbf{Comprehensive multi-domain evaluation:} \systemname{} achieves 77.14\% average linear-probing accuracy across four sensor domains, exceeding the best baseline by 3.56\% and matching supervised PatchTST (76.20\%). On sensor QA, it achieves 68.87\% average accuracy, outperforming all prior sensor-language models on every benchmark, demonstrating the transferability of \systemname{} across sensor configurations and temporal resolutions. 
    \item \textbf{Curated sensor-language pretraining data:} To support sensor-language alignment at scale, we curate a dataset of 600K sensor-caption pairs spanning over 1 billion time points across diverse sensing domains including health, environment, Internet of Things (IoT), energy, and transportation, which we publicly release together with model weights and code to support further research.
\end{itemize}

\section{Related Work}
\paragraph{Time series and sensor foundation models.} 
Sensor waveform signals from wearable devices are time series. Early efforts learn transferable representations via self-supervised pretraining with masked reconstruction~\citep{dong2023simmtm, patchtst} or contrastive learning~\citep{zhang2022tfc}. More recently, general-purpose time series foundation models span domains including electricity, finance, and climate~\citep{ansari2024chronos, chronos2, moirai, goswami2024moment}. While strong on forecasting, these models are optimized for generative objectives and often generalize poorly to classification, which dominates real-world sensor applications.
Recent work also pretrains domain-specific sensor foundation models on large physiological and behavioral datasets~\citep{abbaspourazad2023large, narayanswamy2024scaling, xu2025lsm, qiu2025towards,lens}. PPG-specific~\citep{pillai2024papagei, saha2025pulse} and ECG-specific~\citep{mckeen2025ecg} models gain accuracy and robustness within their target domains. However, these models remain narrow and do not generalize across sensors and tasks.

\paragraph{LLMs for time series and sensor models.} 
Recently, integrating language models into time series and sensor representation learning has emerged as a promising direction~\citep{sensorlm, langer2025opentslm, pillai2025beyond, luo2025normwear, jin2023time, hu2025context}. 
Approaches such as Time-LLM or FSCA~\citep{jin2023time, hu2025context} reprogram time series inputs into the embedding space of pretrained large language models, enabling LLMs to capture high-level semantic structure in temporal data. 
In parallel, cross-modal approaches learn sensor representations through contrastive alignment with natural language. SensorLM~\citep{sensorlm}, for example, aligns wearable sensor streams from Fitbit devices with text using a hierarchical captioning pipeline and large-scale sensor-text pretraining, which improves zero-shot and few-shot recognition, cross-modal retrieval, and semantic grounding on human activity and health tasks.
However, these approaches are largely restricted to specific sensor domains with fixed input lengths, which limits their generalization across diverse sensors and tasks.
In contrast, we propose a unified language-informed sensor pretraining framework that supports heterogeneous sensor domains, enabling strong generalization across tasks ranging from classification to question answering.

\section{\systemname{} Pretraining}
\label{Approach}
\begin{figure*}[t]
    \centering
    \includegraphics[width=\linewidth]{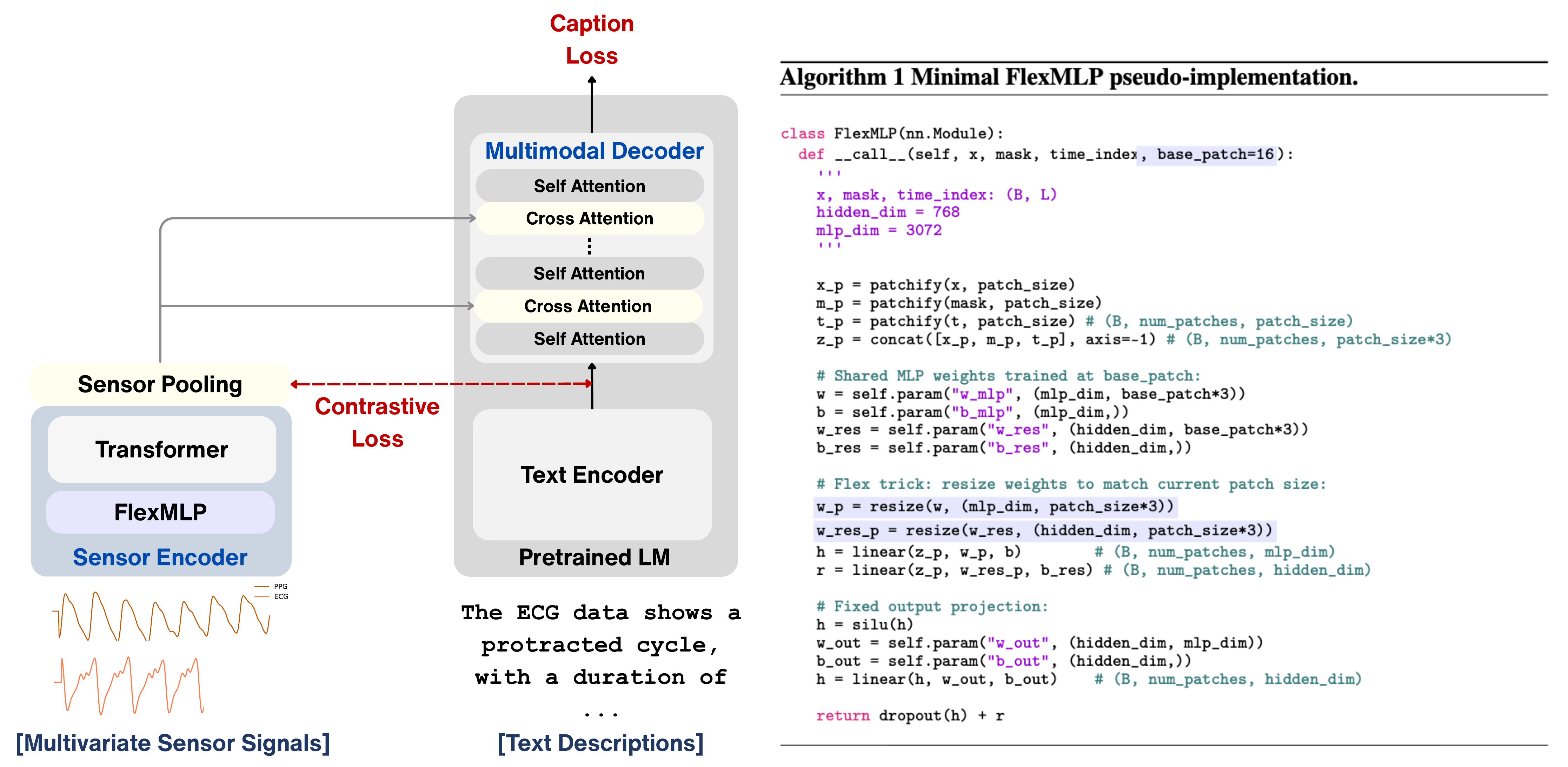}
    \caption{\textbf{SLIP architecture.}
    \textbf{Left:} \systemname{} aligns sensor signals and text with a contrastive captioner: a unified sensor encoder handles varying sensor configurations, and a pretrained LM serves as both text encoder and multimodal decoder.
    \textbf{Right:} Minimal FlexMLP pseudocode. MLP weights learned at \texttt{base\_patch}$=16$ are resized to the runtime \texttt{patch\_size}, so a single encoder handles variable-length inputs. \colorbox{blue!10}{Changes to existing code are highlighted.}
    }
    \vspace{-2mm}
    \label{fig:coca}
\end{figure*}

\textbf{Overview. }
We propose \textbf{SLIP} (Figure~\ref{fig:coca}), a transferable sensor-language framework that learns aligned sensor-text representations from diverse sensor signals $X_s$ and their paired text descriptions $X_t$. Specifically, \systemname{} is trained with a contrastive-captioner framework \citep{coca}, grounding sensor representations in free-form language so that a single pretrained model can be queried, described, or classified through text. With \textit{FlexMLP}, our patch embedder, \systemname{} adjusts its patch size to each input's length and sampling rate, so a single shared backbone encodes all inputs without resampling or retraining. This flexibility lets \systemname{} transfer easily across the heterogeneous healthcare sensing stack (from second-scale wearable streams to day-scale clinical signals) without per-dataset retraining or architectural changes. We describe each component in detail below.

\textbf{Sensor Encoder.}
Behavioral and physiological representations are highly compressible since the patterns are repetitive. To capture this structure, \systemname{} uses a sensor encoder that compresses sensor inputs $X_s$ into compact embeddings $Z_s$, a sequence of continuous vectors analogous to ``tokens''. Each channel is split into patches and embedded by \textit{FlexMLP} (described below). The resulting per-channel patch tokens are concatenated into a single 1D sequence and processed by transformer blocks whose self-attention enables global cross-channel and long-range temporal interactions, yielding the sensor embeddings $Z_s$. Within each block, we apply 2D RoPE~\citep{su2023rope} to preserve the channel-time structure of the flattened sequence, so the model can distinguish each token by both its channel index and its temporal position. 
Following CoCa~\citep{coca}, an attention pooler with $M$ learnable queries cross-attends over $Z_s$, producing a fixed-length output $Z_s'$. The first output token, $z_{\mathrm{cls}}$, serves as the global sensor representation for the contrastive loss, and the remaining $M-1$ tokens are passed as prefix inputs to the multimodal decoder.

\textbf{FlexMLP. }
Existing sensor and time series foundation models use a fixed patch size and resample inputs to a common rate (e.g., 65\,Hz in \citet{luo2025normwear}), which either inflates token counts or discards fine-grained structure; we instead keep each signal at its native rate and adapt the patch size.
FlexMLP is a patch-embedder whose single set of MLP weights is parameterized at a base patch size (i.e., 16) and resized on the fly to any runtime patch size~\citep{beyer2023flexivit}. With no additional parameters or computational overhead, FlexMLP processes long sensor inputs (e.g., EEG) at their native sampling rate by assigning them a larger patch size, bounding the number of tokens per input. Figure~\ref{fig:coca} (right) summarizes a general recipe to flexify an existing MLP patch-embedder~\citep{chronos2}. We provide a detailed analysis and a case study in Appendix~\ref{sec:flexmlp_analysis} to demonstrate the practical value of FlexMLP.

\textbf{Text Encoder and Multimodal Decoder. }
The \emph{Text Encoder} $(X_t \mapsto Z_t)$ produces unimodal text
representations $Z_t$ for contrastive alignment, and the
\emph{Multimodal Decoder}
$(\langle Z_s', Z_t \rangle \mapsto \hat{X}_t)$ fuses these with
the pooled sensor embeddings $Z_s'$ to generate the target text
$\hat{X}_t$ (Figure~\ref{fig:coca}). Rather than train these branches from scratch
(as in CoCa~\citep{coca}), \systemname{} repurposes a single
pretrained language model as the backbone of both: its lower
layers serve as the text encoder, and its upper layers become
the multimodal decoder by inserting a cross-attention sublayer
into each block, so that text tokens can attend to $Z_s'$
during autoregressive decoding. This lets \systemname{} inherit
a strong language prior for both contrastive matching and
free-form sensor description at zero extra pretraining cost.

\textbf{Sensor Language Alignment. }
SLIP is optimized with two objectives jointly: (1) a contrastive loss~\citep{radford2021clip} that aligns global sensor and text embeddings, scoring matched sensor-caption pairs above mismatched ones in the batch; and (2) a captioning loss that trains a multimodal decoder to autoregressively generate the caption, providing a denser supervision signal that captures finer-grained temporal structure, following CoCa \citep{coca}. 

\textbf{Pretraining Dataset Curation. }
Pretraining SLIP requires large-scale paired time series and text data, which is far less available than in vision-language settings. We start from community-released time series corpora~\citep{liu2024utsd, luo2025normwear} that provide diverse signals without aligned text, and we generate multi-level captions at statistical, structural, and semantic levels following the SensorLM recipe~\citep{sensorlm}. To increase pattern diversity, we augment with synthetic time series--text pairs from ChatTS~\citep{chatts}. The resulting set contains over 600K samples and roughly one billion time points spanning energy, environment, health, IoT, nature, transportation, and web domains, with sampling rates from seconds to months and varied sequence lengths. To reduce template repetition, we prompt Qwen2-7B-IT~\citep{qwen2} to generate three paraphrases per caption and randomly sample one during training. As in ChatTS, we sample multivariate and univariate examples in a 2:1 ratio. Corpus statistics are summarized in Table~\ref{tab:category_dist}.

\section{Experimental Setup}
\label{result}

\textbf{Implementation Details. }
\systemname{} features a sensor encoder and a text encoder-decoder, with 402M total parameters (189M trainable). The sensor encoder is a 120M-parameter Transformer~\citep{vaswani2023attentionneed,patchtst} whose patch tokens are pooled via cross-attention with $M=65$ learnable queries: a \emph{classification} (CLS) token whose output serves as the global sensor embedding for computing the contrastive loss, and 64 caption queries that condition the multimodal decoder, following BLIP-2~\citep{li2023blip2}. The text branch is built on Gemma-3-270M~\citep{gemma3}, with the first 12 layers serving as the unimodal text encoder and the final 6 layers turned into a multimodal decoder by inserting a cross-attention layer into each, so that text tokens can attend to sensor representations during autoregressive decoding. Full architectural choices and pretraining details are in Appendix~\ref{implementation}.

\begin{figure*}
    \centering
    \includegraphics[width=1\linewidth]{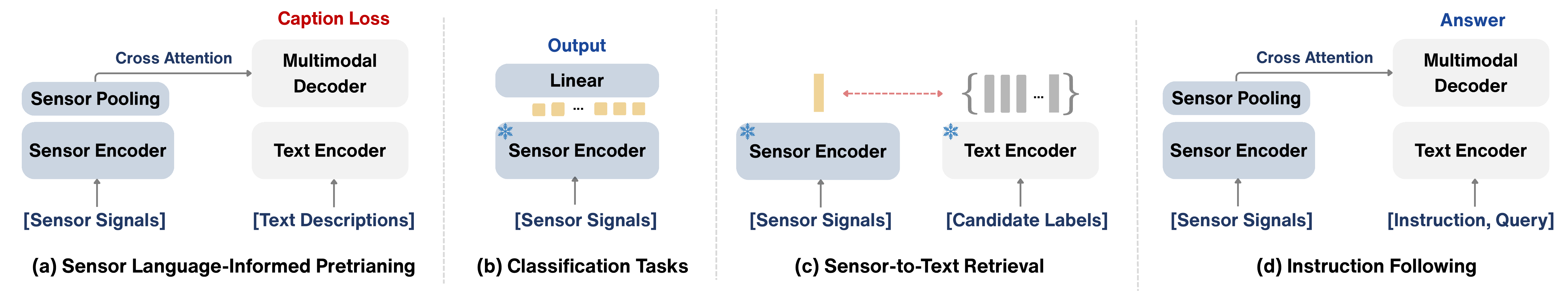}
    \caption{\textbf{Overview of the pretrained \systemname{} and its three downstream tasks:} (a) \textit{pretraining} of \systemname{} via joint sensor-text contrastive matching and caption generation, (b) \textit{classification} via linear probing on frozen sensor encoder features, (c) zero-shot \textit{sensor-text retrieval} via cosine similarity between frozen sensor and text encoder embeddings, and (d) \textit{instruction following} (sensor captioning and question answering) after supervised finetuning.}
    \label{fig:task_demo}
\end{figure*}

\textbf{Downstream Tasks. }
As shown in Figure~\ref{fig:task_demo}, we assess how \systemname{} transfers across three downstream tasks using standard evaluation protocols~\citep{radford2021clip,li2023blip2}:
\begin{itemize}[leftmargin=1.2em,itemsep=2pt,topsep=2pt]
    \item \textit{Classification.} We train a linear classifier on top of the frozen sensor encoder features.
    \item \textit{Retrieval.} Retrieval probes the cross-modal alignment between sensor and text representations in the shared embedding space, with no labeled adaptation at any stage (\textit{i.e.}, zero-shot). We evaluate \emph{sensor-to-text retrieval} only: each class label is converted into a templated sentence, and each sensor sample is assigned the label whose sentence embedding has the highest cosine similarity with its frozen sensor embedding.
    \item \textit{Instruction following.} We finetune \systemname{} to perform both \emph{sensor captioning}, where it generates a natural language description of the signals, and \emph{question answering (QA)}, where it answers questions about the sensor data (e.g., identifying the activity given the accelerometer readings).
\end{itemize}

\textbf{Metrics. } 
We report top-1 accuracy for classification, retrieval, and QA, and evaluate sensor captioning with five text generation metrics: BLEU@4~\citep{bleu}, METEOR~\citep{meteor}, and ROUGE-L~\citep{rouge} measure the lexical $n$-gram overlap with the reference caption, while SBERTSimilarity~\citep{sbert} and BERTScore~\citep{bertscore} capture semantic similarity through contextual embeddings. Higher is better for all metrics.

\textbf{Datasets. } 
For classification and retrieval tasks, we compile 11 datasets across 4 domains: \textit{activity recognition} (WISDM~\citep{wisdm}, UCI-HAR~\citep{ucihar}), \textit{clinical diagnosis} (Stroke, Diabetes, Hypertension~\citep{ppgchina}, Sleep Stage~\citep{sleepedf}, Heart Condition~\citep{ptbxl}), \textit{stress prediction} (WESAD~\citep{wesad}, StudentLife~\citep{studentlife}), and \textit{urban sensing} (Obstacles~\citep{obstacles}, BeijingAQI~\citep{beijingaqi}). All evaluation datasets are held out from our pretraining corpus. Urban sensing serves as an out-of-domain stress test, since our pretraining corpus contains limited in-domain data beyond healthcare. Table~\ref{tab:long} summarizes the sensor modalities and label taxonomies for each dataset. For instruction following, we use five datasets~\citep{langer2025opentslm}: M4 for sensor captioning, and TSQA, HAR-CoT, Sleep-CoT, and ECG-QA-CoT for question answering. All four QA datasets follow the original free-form generation protocol, with TSQA probing basic sensor understanding and the three CoT datasets requiring chain-of-thought reasoning.

\textbf{Baselines.}
Since each task in Figure~\ref{fig:task_demo} is supported by a different set of prior models, we select baselines per task.
(i) For \textit{classification}, we compare against three categories of representation learners: classical ML models (a random forest on catch22 features~\citep{catch22}, SimMTM~\citep{dong2023simmtm}, and TF-C~\citep{zhang2022tfc}), TSFMs (Sundial~\citep{liu2025sundial}, Chronos~\citep{ansari2024chronos}, Chronos-2~\citep{chronos2}, and MOMENT~\citep{goswami2024moment}), and SensorFMs (NormWear~\citep{luo2025normwear} and SensorLM~\citep{sensorlm}).\footnote{As SensorLM is not open-sourced, we reimplement it following \citet{sensorlm} and pretrain it on the same corpus as \systemname{}.} We additionally report a fully supervised PatchTST~\citep{patchtst} as an upper-bound reference.
(ii) For \textit{retrieval}, we only compare against SensorFMs (NormWear and SensorLM), the only families that support sensor-text retrieval.
(iii) For \textit{instruction following} (captioning and QA), we compare against the soft-prompting (SP) and Flamingo variants of OpenTSLM~\citep{langer2025opentslm} and SensorLM, all sharing the same Gemma-3-270M language backbone as \systemname{} for a controlled comparison. 

\section{Results \& Discussion} \label{sec:results}

\begin{table*}[]
\caption{\textbf{Evaluation of the \systemname{} on 11 downstream sensor tasks across 4 domains compared against multiple baselines with linear-probing.} We report the top-1 accuracy on the test set. Best results are in \textbf{bold} and second best are \underline{underlined}.}
\label{tab:cls_ret}
\resizebox{\linewidth}{!}{%
\begin{tabular}{llc|c|cc|c|ccccc|c|cc|c|cc}
\multicolumn{1}{c}{} & \multicolumn{1}{c}{} & &
\multicolumn{3}{c|}{\textbf{\shortstack{Activity\\Recognition}}} &
\multicolumn{6}{c|}{\textbf{\shortstack{Clinical\\Diagnosis}}} &
\multicolumn{3}{c|}{\textbf{\shortstack{Stress\\Prediction}}} &
\multicolumn{3}{c}{\textbf{\shortstack{Urban\\Sensing}}} \\
\multicolumn{1}{c}{\multirow{2}{*}{\textbf{Category}}} &
\multicolumn{1}{c}{\multirow{2}{*}{\textbf{Model}}} &
\rotatebox{90}{\textbf{\# Parameters}} &
\rotatebox{90}{\textbf{Average}} &
\rotatebox{90}{\textbf{WISDM}} &
\rotatebox{90}{\textbf{UCI-HAR}} &
\rotatebox{90}{\textbf{Average}} &
\rotatebox{90}{\textbf{Stroke}} &
\rotatebox{90}{\textbf{Diabetes}} &
\rotatebox{90}{\textbf{Hypertension \quad}} &
\rotatebox{90}{\textbf{Sleep Stag.}} &
\rotatebox{90}{\textbf{Heart Cond.}} &
\rotatebox{90}{\textbf{Average}} &
\rotatebox{90}{\textbf{WESAD}} &
\rotatebox{90}{\textbf{Studentlife}} &
\rotatebox{90}{\textbf{Average}} &
\rotatebox{90}{\textbf{Obstacles}} &
\rotatebox{90}{\textbf{Beijing AQI}} \\[0.8ex] \toprule

\multirow{4}{*}{ML} & \textit{\color{mygray}\shortstack[l]{PatchTST\\(Supervised)}} & \textit{\color{mygray}0.9M} &
\textit{\color{mygray}85.08} & \textit{\color{mygray}85.66} & \textit{\color{mygray}84.49} &
\textit{\color{mygray}76.52} & \textit{\color{mygray}90.91} & \textit{\color{mygray}82.96} & \textit{\color{mygray}48.48} & \textit{\color{mygray}84.29} & \textit{\color{mygray}75.94} &
\textit{\color{mygray}62.20} & \textit{\color{mygray}75.78} & \textit{\color{mygray}48.62} &
\textit{\color{mygray}80.51} & \textit{\color{mygray}84.91} & \textit{\color{mygray}76.11} \\

 & Statistical ML & \textminus &
78.34 & 72.43 & 84.24 &
70.87 & 90.15  & 82.22 & 37.88 & 78.95 & 65.15 &
55.70 & 62.78 & 48.62 &
74.97 & 81.33 & 68.60 \\

 & SimMTM & 0.5M &
58.35 & 44.93 & 71.75 &
62.55 & 89.39 & 82.22 & 38.64 & 46.34 & 56.18 &
62.20 & 75.78 & 48.62 &
71.85 & 76.47 & 67.23 \\

 & TF-C & 12M &
59.60 & 55.39 & 63.81 &
63.66 & \underline{90.91} & \underline{82.96} & 39.39 & 46.97 & 58.06 &
49.58 & 56.95 & 42.20 &
47.98 & 49.87 & 46.08 \\

\midrule
\multirow{4}{*}{TSFM} & Sundial & 128M
& 70.07 & \underline{81.25} & 58.89 &
69.04 & \underline{90.91} & 82.22 & 42.42 & 68.69 & 60.97 &
53.94 & 56.50 & \underline{51.38} &
67.68 & 68.80 & 66.55 \\

 & Chronos & 200M
& 81.07 & 80.30 & 81.84 &
70.85 & 81.06 & 77.78 & 46.97 & 76.68 & \underline{71.76} &
57.50 & 66.37 & 48.62 &
77.02 & 75.19 & \textbf{78.84} \\

 & Chronos2 & 120M
& 77.74 & 75.02 & 80.45 &
71.89 & \underline{90.91} & 77.04 & \underline{47.73} & \underline{82.48} & 61.27 &
60.40 & 73.09 & 47.71 &
65.67 & 81.84 & 49.49 \\

 & MOMENT & 385M
& 71.35 & 69.99 & 72.71 &
71.98 & 82.88 & 78.22 & 45.91 & 80.06 & \textbf{72.85} &
60.01 & 71.75 & 48.26 &
73.76 & \underline{85.68} & 61.84 \\

\midrule
\multirow{3}{*}{SensorFM} & NormWear & 136M
& 74.98 & 70.77 & 79.19 &
\underline{73.00} & \underline{90.91} & 82.22 & 40.15 & \textbf{82.57} & 69.15 &
63.99 & \underline{80.27} & 47.71 &
79.06 & 83.38 & 74.74 \\


 & SensorLM & 114M
& \underline{82.70} & 76.57 & \textbf{88.83} &
71.32 & 85.45 & 75.11 & 44.24 & 80.74 & 71.05 &
\underline{64.08} & 79.55 & 48.61 &
\underline{79.63} & 83.22 & 76.04 \\

\rowcolor{headerblue}
\cellcolor{white} & \textbf{\systemname{}} & 120M
& \textbf{84.31} & \textbf{82.36} & \underline{86.25} &
\textbf{75.79} & \textbf{91.67} & \textbf{84.44} & \textbf{50.00} & 82.04 & 70.79 &
\textbf{68.55} & \textbf{82.96} & \textbf{54.13} &
\textbf{81.96} & \textbf{86.45} & \underline{77.47} \\
\bottomrule
\end{tabular}%
}
\end{table*}
\begin{table*}[t]
\caption{\textbf{Retrieval evaluation on 11 downstream tasks.} We compare \systemname{} with NormWear and SensorLM under the same sensor-text retrieval setting. All results report top-1 accuracy (R@1) on the test set. Best results are in \textbf{bold} and second best are \underline{underlined}.}
\label{tab:zeroshot_result}
\resizebox{\linewidth}{!}{%
\begin{tabular}{l|c|cc|c|ccccc|c|cc|c|cc}
\multicolumn{1}{c}{} &
\multicolumn{3}{c|}{\textbf{\shortstack{Activity\\Classification}}} &
\multicolumn{6}{c|}{\textbf{\shortstack{Clinical\\Diagnosis}}} &
\multicolumn{3}{c|}{\textbf{\shortstack{Stress\\Prediction}}} &
\multicolumn{3}{c}{\textbf{\shortstack{Urban\\Sensing}}} \\
\textbf{Model} &
\rotatebox{90}{\textbf{Average}} &
\rotatebox{90}{\textbf{WISDM}} &
\rotatebox{90}{\textbf{UCI-HAR}} &
\rotatebox{90}{\textbf{Average}} &
\rotatebox{90}{\textbf{Stroke}} &
\rotatebox{90}{\textbf{Diabetes}} &
\rotatebox{90}{\textbf{Hypertension \quad}} &
\rotatebox{90}{\textbf{Sleep Stag.}} &
\rotatebox{90}{\textbf{Heart Cond.}} &
\rotatebox{90}{\textbf{Average}} &
\rotatebox{90}{\textbf{WESAD}} &
\rotatebox{90}{\textbf{Studentlife}} &
\rotatebox{90}{\textbf{Average}} &
\rotatebox{90}{\textbf{Obstacles}} &
\rotatebox{90}{\textbf{Beijing AQI}} \\
\midrule

NormWear &
8.45 & 3.91 & 12.99 &
\textbf{52.52} & \textbf{89.39} & \textbf{82.22} & \textbf{36.36} & \textbf{39.09} & \underline{15.52} &
\underline{29.86} & 16.59 & \textbf{43.12} &
13.35 & 24.30 & 2.39 \\

SensorLM &
\textbf{17.86} & \textbf{9.66} & \underline{26.05} &
41.07 & 45.00 & 67.11 & 28.64 & 15.96 & \textbf{48.64} &
28.60 & \underline{38.12} & 19.08 &
\underline{19.22} & \underline{25.53} & \underline{12.90} \\

\cellcolor{headerblue}\textbf{\systemname{}} &
\cellcolor{headerblue}\underline{17.54} & \cellcolor{headerblue}\underline{7.45} & \cellcolor{headerblue}\textbf{27.62} &
\cellcolor{headerblue}\underline{49.11} & \cellcolor{headerblue}\underline{86.40} & \cellcolor{headerblue}\underline{74.07} & \cellcolor{headerblue}\underline{34.85} & \cellcolor{headerblue}\underline{34.83} & \cellcolor{headerblue}15.39 &
\cellcolor{headerblue}\textbf{49.12} & \cellcolor{headerblue}\textbf{56.95} & \cellcolor{headerblue}\underline{41.28} &
\cellcolor{headerblue}\textbf{27.08} & \cellcolor{headerblue}\textbf{29.92} & \cellcolor{headerblue}\textbf{24.23} \\

\bottomrule
\end{tabular}%
}
\end{table*}

\subsection{Downstream Classification}
Linear probing on frozen features is a standard protocol for evaluating the quality of pretrained representations~\citep{radford2021clip}. We adopt this protocol to compare representations learned with language supervision against those learned with predictive SSL objectives.
Across 11 datasets spanning 4 domains (activity recognition, clinical diagnosis, stress prediction, and urban sensing), \systemname{} achieves the highest average accuracy in every domain (Table~\ref{tab:cls_ret}).
The margin is largest on stress prediction ($+$4.56\% over NormWear, the strongest predictive SSL baseline), where labels correspond to behavioral and physiological states (e.g., elevated heart rate, reduced movement, phone-usage patterns) that the pretraining captions describe explicitly. Language supervision therefore aligns the embedding space directly with the factors that define these labels, making them linearly separable in the frozen features.
Conversely, \systemname{} slightly underperforms predictive SSL baselines on Sleep Staging ($-$0.53\% vs.\ NormWear) and Heart Condition ($-$2.06\% vs.\ MOMENT). These tasks depend on fine-grained waveform morphology (e.g., sleep spindles, QRS-complex variations), which captions abstract away but signal-level objectives preserve.
More broadly, both language-supervised models outperform all TSFMs pretrained with reconstruction objectives (77.14\% average accuracy for \systemname{} and 73.58\% for SensorLM). This result suggests that language supervision yields more transferable representations than raw-signal prediction. Because captions are invariant to sensor noise and other low-level nuisance factors, the language-supervised objective encourages the encoder to discard these factors and retain semantic structure.
The learned representations are expressive enough that a linear probe on frozen \systemname{} features performs comparably to a fully supervised PatchTST trained end-to-end on each task.
Compared with SensorLM, a controlled baseline trained on the same corpus with the same objectives, \systemname{} improves average accuracy by 3.56\% and outperforms it on most datasets, demonstrating the effectiveness of its architectural design.
A pairwise Wilcoxon test~\citep{wilcoxon1945individual} confirms that \systemname{} differs significantly from every baseline at $p<0.05$ (Appendix~\S\ref{sec:stats_sig}); accuracy over 5 random seeds and F1 results are provided in Tables~\ref{tab:kfold_res} and~\ref{tab:f1_results}.

\subsection{Sensor-to-Text Retrieval}
Although a joint sensor-language embedding space offers the flexibility to classify new sensor tasks from text descriptions alone, this setting is far harder than linear probing, since no classifier is fit to labeled data and accuracy depends entirely on the pretrained sensor-text alignment.
NormWear addresses this by training an additional temporal fusion module that aligns its self-supervised sensor embeddings with text embeddings for zero-shot inference, and we therefore evaluate NormWear equipped with this module as the baseline in this setting.
As Table~\ref{tab:zeroshot_result} shows, \systemname{} achieves the highest average accuracy across all 11 datasets (39.36\% vs.\ 33.26\% for NormWear and 30.61\% for SensorLM). The largest gain is on stress prediction, whose labels (e.g., stress, amusement) denote abstract affective states rather than directly observable signal patterns. \systemname{} inherits a text space that already encodes the meaning of such labels from its pretrained LM initialization, allowing contrastive pretraining to ground sensor patterns directly in these abstract concepts.
On clinical diagnosis, NormWear outperforms \systemname{} slightly (52.52\% vs.\ 49.11\%) because its text encoder, ClinicalTinyLlama, is pretrained on clinical text and encodes concepts such as \emph{hyperglycemia} and \emph{autonomic neuropathy} that are rare in Gemma-3-270M's general-domain corpus. Overall, these results suggest that repurposing a pretrained LM, rather than learning text representations from sensor captions alone, carries broad language knowledge into the shared space and benefits sensor representation learning.

\subsection{Instruction Following}
\paragraph{Sensor Captioning.}
Unlike TSFMs, \systemname{} supports open-vocabulary generation, allowing it to answer arbitrary questions about sensor signals. We first evaluate this capability on sensor captioning, where the model describes the input signal in natural language (Figure~\ref{fig:m4qa_demo}). As shown in Table~\ref{tab:sft_res}, \systemname{} outperforms both OpenTSLM variants across all metrics, from surface overlap (BLEU@4, ROUGE-L, METEOR) to semantic similarity (SBERTSimilarity, BERTScore). Moreover, we note that \systemname{} performs on par with SensorLM, another language-aligned model, suggesting that language alignment provides a strong starting point for finetuning.

\paragraph{Sensor Question Answering.}
Beyond describing a signal, question answering requires the model to reason over it to reach a specific answer (Figures~\ref{fig:har_cot_demo}, \ref{fig:sleep_cot_demo}, and~\ref{fig:ecg_failure}). As shown in Table~\ref{tab:sft_res}, in a controlled setting where all models have roughly the same number of parameters (120M sensor encoder, 270M language model), \systemname{} achieves the best accuracy on all four benchmarks, outperforming OpenTSLM and SensorLM.
Specifically, the largest gain over OpenTSLM-Flamingo appears on TSQA ($+$21.63\%), which mainly tests global signal patterns, suggesting that the contrastive objective captures high-level semantics while abstracting away low-level noise.
Compared with SensorLM, \systemname{} performs significantly better on ECG-QA-CoT ($+$6.89\%), which requires fine-grained waveform reasoning. As in retrieval, this result again suggests that initializing from a pretrained LM during pretraining helps the sensor encoder learn more fine-grained representations.
Overall, the consistent margin across all four benchmarks suggests that language-aligned sensor representations provide more informative conditioning for the language model.
We further compare against large TSLMs (i.e., ChatTS-8B and ITFormer-0.5B) in Table~\ref{tab:sft_res_full}. Despite \systemname{}'s much smaller language backbone, it outperforms both on all captioning metrics and most QA benchmarks.
\begin{table}[t]
\centering
\caption{\textbf{Sensor captioning and question answering results.} All models share the same Gemma-3-270M language backbone for a controlled comparison. Captioning is evaluated on the M4 test set with BLEU@4, METEOR, ROUGE-L, SBERTSim, and BERTScore. TSQA, HAR-CoT, Sleep-CoT, and ECG-QA-CoT share the same QA evaluation protocol, reporting top-1 accuracy. Higher is better for all metrics. Best results are in \textbf{bold} and second best are \underline{underlined}.}
\label{tab:sft_res}
\resizebox{\linewidth}{!}{%
\begin{tabular}{l|ccccc|cccc}
\toprule
\multirow{2}{*}{\textbf{Model}} &
\multicolumn{5}{c|}{\textbf{Sensor Captioning (M4)}} &
\textbf{TSQA} & \textbf{HAR-CoT} & \textbf{Sleep-CoT} & \textbf{ECG-QA-CoT} \\
\cmidrule(lr){2-6} \cmidrule(lr){7-10}
 &
\textbf{BLEU@4} & \textbf{METEOR} & \textbf{ROUGE-L} & \textbf{SBERTSim} & \textbf{BERTScore} &
\textbf{Accuracy} & \textbf{Accuracy} & \textbf{Accuracy} & \textbf{Accuracy} \\
\midrule
OpenTSLM-SP &
0.0026 & 0.0456 & 0.0255 & 0.1551 & 0.7250 &
26.79 & 0.55 & 5.91 & 1.11 \\
OpenTSLM-Flamingo &
0.1141 & 0.3210 & 0.2894 & 0.7990 & 0.8858 &
78.12 & \underline{63.43} & 68.49 & \underline{35.50} \\
SensorLM &
\underline{0.2008} & \textbf{0.4151} & \underline{0.3595} & \textbf{0.9014} & \underline{0.9024} &
\underline{95.73} & 60.80 & \underline{73.12} & 30.29 \\
\rowcolor{headerblue}
\textbf{\systemname{}} &
\textbf{0.2046} & \underline{0.4027} & \textbf{0.3658} & \underline{0.8977} & \textbf{0.9077} &
\textbf{99.75} & \textbf{64.35} & \textbf{74.19} & \textbf{37.18} \\
\bottomrule
\end{tabular}%
}
\end{table}

\subsection{Ablation Studies}

\label{ablation}


In this section, we conduct an extensive ablation study on \systemname{} to understand how each architectural choice and pretraining objective affects the learned representations and downstream performance. All models follow the same evaluation protocol as in Section~\ref{result}, and we report only the average performance for each task. Full results are provided in Appendix~\ref{full_result}.

\textbf{Training Objectives. }
\systemname{} is jointly optimized with a contrastive loss and a captioning loss. To understand what each loss contributes, we train variants with only one of the two (Table~\ref{tab:ablations}(a)). Removing the captioning loss (\emph{Contrastive-only}) barely affects retrieval ($-2.46$) but causes a large drop in QA ($-15.81$). Removing the contrastive loss (\emph{Caption-only}) has the opposite effect: QA drops moderately ($-6.02$), while retrieval drops sharply ($-14.24$). These results suggest that the two objectives are complementary. The contrastive loss aligns a global sensor embedding with text, which supports zero-shot retrieval, whereas the captioning loss provides token-level supervision that supports generating answers but does not necessarily produce a well-structured joint embedding space. Appendix~\ref{sec:lunif} supports this with an alignment and uniformity analysis of the embedding space. 

\textbf{FlexMLP. }
Sequence lengths vary by orders of magnitude across datasets. FlexMLP therefore shares one set of patch-embedding weights across patch sizes, so a single pretrained encoder can tokenize each dataset at a suitable patch size without retraining. As Table~\ref{tab:ablations}~(b) shows, FlexMLP improves performance on every task, with the largest gain on retrieval ($+4.57$). Details of the patch-size heuristic are in Appendix~\ref{sec:ps_sel}.


\textbf{Cross-sensor Learning. }
Cross-channel interactions are typically modeled with either full self-attention over all channel-time tokens or memory-efficient group attention~\citep{chronos2}, which restricts attention to within channel groups. In Table~\ref{tab:ablations}(c), the two perform comparably on perception and retrieval, while full attention is slightly better on QA ($+2.40$). We therefore keep full attention as the default and consider group attention a practical alternative under tight memory budgets.

\subsection{Further Analysis and Sensitivity Studies}

\textbf{Sensor and text encoder size. }
By default, \systemname{} pairs a 120M sensor encoder with a text encoder of comparable scale (Gemma-3-270M). We investigate whether a larger text encoder helps representation learning, as it encodes more real-world knowledge and may provide a better representation space for the sensor encoder to align with. We train \systemname{} with Llama-3.2-1B as the text encoder (Table~\ref{tab:ablations}~(e)), and surprisingly it underperforms the default Gemma-3-270M on sensor perception and retrieval, with only a marginal gain on QA ($+1.12$). This result suggests that it is more beneficial for the sensor and text encoders to have roughly the same number of parameters. To further validate this, we train a smaller sensor encoder (40M) with Gemma-3-270M and observe significant performance degradation on retrieval and QA, suggesting that the sensor features remain linearly separable but are poorly aligned with the text representation space. Together, these findings indicate that when one branch dominates in capacity, the contrastive objective provides weaker gradients to the smaller branch. We leave mitigations (\textit{e.g.}, branch-specific learning rates or LoRA on the text branch) to future work.

\textbf{Partial finetuning of text encoder. }
We explore whether freezing the text encoder allows the sensor encoder to learn a direct mapping into the text representation space, effectively translating sensor input into language. From Table~\ref{tab:ablations}~(f), we observe that doing so significantly degrades performance compared to our default \systemname{}, especially on QA ($-8.32$). As shown in Figure~\ref{fig:contrastive_analysis}, sensor representations collapse toward a limited region of the feature space, leading to consistent degradation across downstream tasks. This result suggests that a frozen text encoder constrains the representation space too rigidly, forcing the sensor encoder to map all inputs into a narrow region rather than learning distributed representations. These results suggest that jointly finetuning both branches helps the two modalities co-adapt into a better-structured joint representation space.

\begin{table*}[t]
    \caption{\textbf{Ablation study results}. The default setting adopted by SLIP is marked in \colorbox{headerblue}{blue}. We calculate $\pm$ $\Delta$ within each group of ablations in comparison with the default setting. }
    \label{tab:ablations}
    \centering
    
    \begin{minipage}[t]{0.525\textwidth}
        \resizebox{\linewidth}{!}{%
        \begin{tabular}{r r l r l r l}
            \toprule
             & \multicolumn{2}{c}{\begin{tabular}{@{}c@{}}\textbf{Sensor Perception}\\(Accuracy)\end{tabular}} & \multicolumn{2}{c}{\begin{tabular}{@{}c@{}}\textbf{Retrieval}\\(Accuracy)\end{tabular}} & \multicolumn{2}{c}{\begin{tabular}{@{}c@{}}{\textbf{QA}}\\(Accuracy)\end{tabular}} \\
            \midrule
            \multicolumn{7}{l}{\textit{\color{mygray}(a) Training Objectives}} \\
            \rowcolor{headerblue}
            SLIP pretraining & 77.14 & & 39.36 & & 68.87 & \\
            Caption-only & 74.30 & \negval{-2.84} & 25.12 & \negval{-14.24} & 62.85 & \negval{-6.02} \\
            Contrastive-only & 74.77 & \negval{-2.37} & 36.90 & \negval{-2.46} & 53.06 & \negval{-15.81} \\
            \bottomrule
            \toprule

            \multicolumn{7}{l}{\textit{\color{mygray} (b) FlexMLP}} \\
            \rowcolor{headerblue}
            w/ FlexMLP & 77.14 & & 39.36 & & 68.87 & \\
            w/o FlexMLP (\textit{patch size} = 16) & 74.16 & \negval{-2.98} & 34.79 & \negval{-4.57} & 66.88 & \negval{-1.99} \\
            \bottomrule \toprule
            
            \multicolumn{7}{l}{\textit{\color{mygray} (c) Cross-Sensor Learning }} \\
            \rowcolor{headerblue}
            Self-Attention w/ 2D RoPE& 77.14 &  & 39.36 &  & 68.87 &  \\
            Group Attention & 77.04 & \negval{-0.10} & 39.21 & \negval{-0.15} & 66.47 & \negval{-2.40} \\
            \bottomrule
            
        \end{tabular}%
        }
    \end{minipage}
    \hfill
    \begin{minipage}[t]{0.455\textwidth}
        \resizebox{\linewidth}{!}{%
        \begin{tabular}{r r l r l r l}
            \toprule
             & \multicolumn{2}{c}{\begin{tabular}{@{}c@{}}\textbf{Sensor Perception}\\(Accuracy)\end{tabular}} & \multicolumn{2}{c}{\begin{tabular}{@{}c@{}}\textbf{Retrieval}\\(Accuracy)\end{tabular}} & \multicolumn{2}{c}{\begin{tabular}{@{}c@{}}{\textbf{QA}}\\(Accuracy)\end{tabular}} \\
            \midrule
            \multicolumn{7}{l}{\textit{\color{mygray} (d) Sensor Encoder Parameter Size }} \\
            \rowcolor{headerblue}
            SLIP\textsubscript{Base} (120M) & 77.14 & & 39.36 & & 68.87 & \\
            SLIP\textsubscript{Small} (40M) & 74.84 & \negval{-2.30} & 26.71 & \negval{-12.65} & 59.93 & \negval{-8.94} \\
            \bottomrule \toprule

            \multicolumn{7}{l}{\textit{\color{mygray} (e) Language Model }} \\
            \rowcolor{headerblue}
            Gemma-3-270M & 77.14 & & 39.36 & & 68.87 & \\
            Llama3.2-1B & 75.03 & \negval{-2.11} & 36.53 & \negval{-2.83} & 69.99 & \posval{+1.12} \\
            \bottomrule \toprule
            
            \multicolumn{7}{l}{\textit{\color{mygray} (f) Partial finetuning of text-encoder}} \\
            \rowcolor{headerblue}
            Finetune 4-layer & 77.14 &  & 39.36 &  & 68.87 & \\
            Freeze& 73.56 & \negval{-3.58} & 35.68 & \negval{-3.68} & 60.55 & \negval{-8.32} \\
            \bottomrule
            
        \end{tabular}%
        }
    \end{minipage}
\vspace{-0.42cm}
\end{table*}

\section{Discussion and Conclusion}
\label{sec:conclusion}

\textbf{Limitations and future work.}\label{sec:limitation}
Two challenges remain open for sensor-language pretraining models. First, compared to vision-language models that rely on short captions, sensor-language models must condition on longer free-form descriptions to express signal semantics, increasing context length and computational cost. Scalability could be improved through selective decoding or adaptive context compression. Second, principled analyses of hallucination, output faithfulness, and temporal attribution remain largely absent for sensor-language generation. Developing such attribution and faithfulness methods remains an important direction for future work.

\textbf{Conclusion and broader impact.}
This paper introduces \systemname{}, the first framework for learning transferable sensor representations from natural language supervision, trained with contrastive alignment and sensor-conditioned captioning in a single stage. By repurposing a pretrained decoder-only language model into a multimodal encoder-decoder via cross-attention, \systemname{} unifies perception, retrieval, and generation in one architecture. From a single checkpoint, \systemname{} achieves state-of-the-art performance across diverse sensor domains. It also adapts to complex sensor question answering with little finetuning. Unlike prior sensor encoders such as MOMENT, \systemname{} is, to our knowledge, the first that is both language-aligned and cross-modal across sensor types. By mapping heterogeneous sensor signals into a shared language-aligned space, \systemname{} can serve as a building block for future sensor-language models, and lower the barrier to downstream systems for health monitoring, activity understanding, accessibility, and human-centered AI. We hope this work motivates new directions for large sensor-language foundation models.

\section*{Acknowledgments}
The authors acknowledge support for this research from \textbf{Evergreen: A Generative AI and Behavioral Sensing Digital Ecosystem to Promote Student Wellness and Flourishing.} This work is made possible through philanthropic gifts to Dartmouth College dedicated to advancing AI-supported well-being and flourishing of college students.

\bibliography{reference}
\bibliographystyle{iclr2027_conference}


\appendix
\clearpage
\section{Appendix}
\subsection{Datasets}
We thank the authors of prior work who collected and open-sourced the datasets used in this work, and we summarize them again below for the convenience of future researchers. We provide the details of the non-overlapping pretraining and downstream datasets below.
\label{sec:dataset}
\subsubsection{Pretraining Dataset}
Inspired by SensorLM \citep{sensorlm}, we automatically generate hierarchical captions (i.e., statistical, structural, semantic) of each multivariate sensor signal. The domain distribution of this sensor-paired dataset is shown in Table~\ref{tab:category_dist}. We will open-source this large dataset to support future research on sensor-language models, and we will also release the caption generation pipeline for creating large-scale sensor-language datasets.

\label{sec:pretrain_ds}
\begin{table}[h]
\centering
\caption{Category distribution.}
\label{tab:category_dist}
\resizebox{\linewidth}{!}{%
\begin{tabular}{lcccccccc|c}
\toprule
 & Health & Web & Synthetic & Nature & Energy & IoT & Environment & Transport & Total \\
\midrule
\# of Samples & 285054 & 145063 & 105085 & 69518 & 5846 & 5508 & 2368 & 66 & 618508 \\
Percent (\%) & 46.09 & 23.45 & 16.99 & 11.24 & 0.95 & 0.89 & 0.38 & 0.01 & 100.00 \\
Sources
& \srcthree{\citep{liu2024utsd}}{\citep{luo2025normwear}}{\citep{capture24}}
& \srcone{\citep{liu2024utsd}}
& \srcone{\citep{chatts}}
& \srcone{\citep{liu2024utsd}}
& \srcone{\citep{liu2024utsd}}
& \srcone{\citep{liu2024utsd}}
& \srcone{\citep{liu2024utsd}}
& \srcone{\citep{liu2024utsd}}
& \\
\bottomrule
\end{tabular}%
}
\end{table}

\subsubsection{Downstream Dataset}
\label{sec:downstream_ds}

\begingroup
\footnotesize
\begin{longtable}{@{} l L{2cm} r c c L{2cm} @{}}
\caption{Evaluation dataset details.} \label{tab:long} \\
\toprule

Dataset &
Sensor &
\shortstack[l]{\# Samples \\ (Train / Test)} &
Freq. &
\# Cls &
Label Names \\
\midrule
\endfirsthead
\toprule

Dataset &
Sensor &
\shortstack[l]{\# Samples \\ (Train / Test)} &
Freq. &
\# Cls &
Label Names \\
\midrule
\endhead

WISDM \citep{wisdm} &
Accelerometer X, Y, Z &
22396 / 5600 &
30Hz &
18 &
Catch, Chips, Clap, Dribble, Drink, Fold, Jog, Kick, Pasta, 
Sandwich, Sit, Soup, Stair, Stand, Teeth, Type, Walk, Write \\
\midrule

UCI-HAR \citep{ucihar} &
Accelerometer \newline X, Y, Z \newline
Gyroscope \newline X, Y, Z &
1847 / 793 &
50Hz &
7 &
Lay, Sit, Stand, Walk,
Walking Upstairs,
Walking Downstairs,
Transition \\
\midrule

\shortstack[l]{PPG-CVA \\ (Stroke) \citep{ppgchina}} &
PPG &
525 / 132 &
65Hz &
2 &
Normal, Stroke \\ 
\midrule

\shortstack[l]{PPG-DM \\ (Diabetes) \citep{ppgchina}} &
PPG &
522 / 135 &
65Hz &
2 &
Normal, Diabetes \\
\midrule

\shortstack[l]{PPG-HTN \\ (Hypertension) \citep{ppgchina}} &
PPG &
525 / 132 &
65Hz &
4 &
Normal, Pre-hypertension, Stage-1, Stage-2 \\
\midrule

\shortstack[l]{Sleep Stage \\ (Sleep stage) \citep{sleepedf}} &
EEG-Fpz-Cz, EEG-Pz-Oz &
33599 / 8709 &
100Hz &
5 &
Light spindle, Light, Deep, REM, Wake \\
\midrule

\shortstack[l]{PTB-XL \\ (Heart cond.)} \citep{ptbxl} &
12-lead ECG &
11320 / 1650 &
100Hz &
5 &
Normal ECG, Myocardial Infarction, 	ST/T Change, Conduction Disturbance, Hypertrophy\\
\midrule

WESAD \citep{wesad}&
Chest Acc. X, Y, Z \newline
Chest ECG, EMG, EDA, Temp, Resp, \newline
Wrist Acc. X, Y, Z, \newline
Wrist BVP, EDA &
882 / 223 &
700Hz &
3 &
Neutral, Amusement, Stress \\
\midrule

StudentLife \citep{studentlife} &
Activity,
Audio,
Conversation,
Phone Charge,
Phone Lock,
Time to deadline,
Day of the week,
Exam period,
Sleep rating,
Sleep duration &
1074 / 109 &
Minute &
3 &
Normal, \newline Medium Stress, \newline High Stress \\
\midrule


AsphaltObstacles \citep{obstacles} &
Acceleration magnitude &
390 / 391 &
100Hz &
4 &
Raised crosswalk, Raised markers, \newline 
Speed bump, \newline 
Vertical patch \\
\midrule

Beijing AQI \citep{beijingaqi}&
dew-point temperature,
windspeed,
PM25,
PM10,
NO2,
SO2,
CO &
1168 / 293 &
Hour &
5 &
Good, Moderate, Unhealthy for sensitive groups, Unhealthy, Hazardous \\

\bottomrule
\end{longtable}
\endgroup
\clearpage
\subsection{Pretraining Dataset Scaling}
Table~\ref{tab:pretrain_size} illustrates the scaling behavior of SLIP across different data regimes. Consistent with the scaling laws observed in the original CLIP work, our results demonstrate that retrieval performance benefits significantly more from large-scale data than classification. While classification accuracy remains relatively stable—dropping only 2.88\% when scaling down from 1.7B to 0.3B samples—retrieval performance collapses by 17.72\%. This follows the intuition from \citet{radford2021clip} that while coarse semantic categories can be learned from limited samples, the construction of a high-fidelity, shared latent space requires the ``density'' of a massive dataset. Large-scale pretraining is essential for the model to learn the fine-grained nuances necessary to distinguish between similar sensor-text pairs in a global retrieval space, suggesting SLIP's greater potential as datasets continue to grow.
\begin{table}[h]
\centering
\caption{\textbf{Effect of Pretraining Dataset Size}}
\label{tab:pretrain_size}
\resizebox{0.6\linewidth}{!}{%
\begin{tabular}{lcccccc}
\toprule
\textbf{Model} & \textbf{Cls. Avg} & $\Delta$ & \textbf{Retrieval} & $\Delta$ & \textbf{QA} & $\Delta$ \\
\midrule
\rowcolor{headerblue}
Ours (1.7 B) & 77.14 &  & 39.36 &  & 68.87 &  \\
0.8 B & 75.23 & \negval{-1.91} & 27.09 & \negval{-12.27} & 60.18 & \negval{-8.69} \\
0.3 B & 74.26 & \negval{-2.88} & 21.64 & \negval{-17.72} & 62.95 & \negval{-5.92} \\
\bottomrule
\end{tabular}
}
\end{table}
\clearpage
\subsection{Implementation Details}
\label{implementation}

\subsubsection{Architecture}
\label{sec:architecture}

\paragraph{Sensor encoder.} The sensor encoder is a standard Transformer backbone with 120M parameters and hidden dimension 768~\citep{vaswani2023attentionneed,patchtst}. Following \citet{liu2025sundial}, we adopt Pre-LN and FlashAttention for training stability and efficiency, and use 2D RoPE~\citep{su2023rope} to preserve the channel-time structure across the flattened token sequence. Instead of the conventional linear projection from input signal patches to hidden dimension, we use FlexMLP with a base patch size of 16 to support variable patch sizes for different sampling rates (see Appendix~\ref{sec:ps_sel} for the patch size heuristic). The resulting tokens are pooled via cross-attention with $M=65$ learnable queries: one CLS query for contrastive matching, and 64 caption queries that condition the multimodal decoder, corresponding to an $8\times$ temporal compression as in BLIP-2~\citep{li2023blip2}. During pretraining, we cap each example at 6 channels and 384 sensor patches to keep training efficient and enable larger contrastive batch sizes.

\paragraph{Text encoder-decoder.} The text encoder is initialized from the first 12 layers of Gemma-3-270M~\citep{gemma3}, placing its parameter budget in the same range as text encoders commonly used in CoCa, while the multimodal decoder is initialized from the final 6 layers of the same model. To condition text generation on sensor representations, we insert a cross-attention layer into each of the final six unimodal layers, transforming them into a multimodal decoder in which text tokens can attend to sensor encoder outputs during autoregressive decoding. For efficiency and training stability, we unfreeze only the last 4 layers of the unimodal text encoder, yielding 22M trainable parameters in the encoder. The effect of freezing the entire text encoder is analyzed in Section~\ref{ablation}. We reuse the Gemma-3-270M tokenizer and token embedding, and cap the input at 512 query tokens, padding shorter queries with [PAD]. After adding cross-attention, the decoder is trained with full parameters (43M trainable). Overall, this branch closely follows the Gemma-3-270M architecture and introduces only 10M additional parameters via cross-attention.

\subsubsection{Patch Size Heuristic}
\label{sec:ps_sel}
Our pretraining corpus mixes sampling rates from monthly to 700Hz and sequence lengths that differ by orders of magnitude (Table~\ref{tab:category_dist}). A single patch size would turn this diversity into padding: short low-rate series would be padded out to the length of long high-rate ones, and most of the compute in a batch would be spent on padding patches. The patch size heuristic is what lets us train on the whole corpus jointly. It assigns a larger patch to longer or lower-rate series, so the token count per sample stays roughly constant across datasets and every token carries real signal. We define the heuristic by frequency, following \citet{moirai}, and widen its range of window sizes to cover our corpus:

\begin{itemize}
    \item \textbf{Daily:} 16, 32
    \item \textbf{Hourly:} 6, 8, 16, 24, 32, 64
    \item \textbf{Minute-level:} 16, 24, 128
    \item \textbf{Second-level:} 4, 6, 8, 12, 16, 20, 25, 32, 64, 128
\end{itemize}

Because the heuristic assigns different patch sizes across frequencies, a single set of patch embedding weights is shared across patch sizes and resized by interpolation at evaluation time. The patch size is thus a choice made after pretraining: a user can trade token count for temporal granularity according to the downstream task, without retraining the encoder. We analyze this behavior in Appendix~\ref{sec:flexmlp_analysis}.


\subsubsection{Pretraining Implementation Details}
We pretrain for 40 epochs with batch size 72 on 4 NVIDIA H200 GPUs using AdamW \citep{adamw} (weight decay = 0.05). The learning rate is warmed up to $2e^{-4}$ over 80k iterations and cosine-decayed to $1e^{-7}$. Sensor augmentations are disabled during pretraining to preserve sensor-text alignment, but standard augmentations (jittering, scaling, time flipping) are applied during supervised finetuning \citep{sensorlm}. Additional implementation details are in the codebase. 

\clearpage
\subsubsection{Downstream Evaluation Details}
Following the standard transfer learning evaluation protocol from prior work \citep{he2021maskedautoencodersscalablevision}, we did not perform hyperparameter search (e.g., learning rate or weight decay). All models were frozen, and only a randomly initialized linear classifier was trained for 50 epochs with 5 warmup epochs using the AdamW optimizer, a base learning rate of 0.01, and a weight decay of 0.05. 
Supervised finetuning hyperparameter details are shown in Table~\ref{tab:hyperparams}.

\begin{table*}[h]
\centering
\caption{\textbf{Hyperparameters used in the supervised finetuning experiment.}}
\label{tab:hyperparams}
\resizebox{0.8\linewidth}{!}{%
\begin{tabular}{lccccc}
\toprule
\textbf{Hyperparameter} &
\textbf{TSQA} &
\textbf{HAR-CoT} &
\textbf{Sleep-CoT} &
\textbf{ECG-QA-CoT} &
\textbf{M4 Caption} \\
\midrule
Optimizer & \multicolumn{5}{c}{AdamW \citep{adamw}} \\
Gradient clip & \multicolumn{5}{c}{1.0} \\
LR decay schedule & \multicolumn{5}{c}{Cosine Schedule Decaying to $1e^{-7}$} \\
Train Epoch & 4 & 4 & 10 & 4 & 4 \\
Train batch size & 64 & 64 & 32 & 32 & 64 \\
Warm up epochs & 1 & 1 & 1 & 1 & 1 \\
Weight decay rate & 0.05 & 0.05 & 0.05 & 0.05 & 0.05 \\
\bottomrule
\end{tabular}%
}
\end{table*}

\subsection{Statistical Comparisons}
\label{sec:stats_sig}

Following \citet{stattest}, we first run a Friedman omnibus test on the linear-probing accuracies of all 10 models across the 11 datasets in Table~\ref{tab:cls_ret}, excluding the supervised PatchTST reference. The test gives $\chi^2_F = 32.7$ with $p = 1.5\times 10^{-4}$, firmly rejecting the null that all models perform identically. We then compare SLIP against each baseline using a two-sided paired Wilcoxon signed-rank test over the 11 datasets, and report the results in Table~\ref{tab:slip_holm_pvalues}. SLIP attains the best average rank ($1.64$) and improves over every baseline at $p \le 0.01$, which remains significant after Holm correction for the 9 comparisons (adjusted $p \le 0.015$). 

For the zero-shot setting we apply the same test against both retrieval baselines. SLIP outperforms SensorLM on 9 of the 11 datasets with a median margin of $6.96\%$ ($W = 12$, $p = 0.067$), while it is statistically indistinguishable from NormWear ($W = 24$, $p = 0.46$, winning on 5 of 11 datasets). The latter is consistent with NormWear's stronger alignment on the diagnosis tasks. We do not report statistical tests for the supervised finetuning experiment because there are fewer than five datasets, following the guidance in~\citet{stattest}.

\begin{table}[h]
\centering
\caption{\textbf{Wilcoxon signed-rank tests of SLIP against each baseline.} Average ranks are over the 11 datasets, lower is better. $^{*}p<0.1$, $^{**}p<0.05$, n.s.\ = not significant.}
\label{tab:slip_holm_pvalues}
\setlength{\tabcolsep}{10pt}
\resizebox{0.5\linewidth}{!}{%
\begin{tabular}{l|lccc}
\toprule
\textbf{Eval} & \textbf{Model} & \textbf{Avg.\ rank} & \textbf{$p$} & \textbf{sig} \\
\midrule
\multirow{10}{*}{LP}
& \textbf{SLIP (ours)} & \textbf{1.64} & \textminus{} & \textminus{} \\
& SensorLM     & 4.64 & 0.0098 & ** \\
& NormWear     & 4.68 & 0.0020 & ** \\
& Chronos      & 5.18 & 0.0049 & ** \\
& Chronos2     & 5.55 & 0.0020 & ** \\
& MOMENT       & 5.82 & 0.0049 & ** \\
& Stat Feat    & 5.86 & 0.0010 & ** \\
& Sundial      & 6.36 & 0.0010 & ** \\
& SimMTM       & 7.23 & 0.0010 & ** \\
& TFC          & 8.05 & 0.0010 & ** \\
\midrule
\multirow{3}{*}{ZS}
& \textbf{SLIP (ours)} & \textbf{1.73} & \textminus{} & \textminus{} \\
& NormWear     & 2.00 & 0.4648 & n.s. \\
& SensorLM     & 2.27 & 0.0674 & * \\
\bottomrule
\end{tabular}%
}
\end{table}

\clearpage
\subsection{FlexMLP}
\label{sec:flexmlp_analysis}
FlexMLP equalizes sequence length across sampling rates by varying the patch size. The same equalization can be reached by resampling all data to one rate instead, as in NormWear (65Hz). Resampling shortens the sequence but also discards the high-frequency content of a signal before the model sees it, and that content often carries the label. Table~\ref{tab:aasm_bands} shows this on Sleep Stage: sleep spindles are removed at 25Hz, and every rhythm above the slow waves is removed at 5Hz. Enlarging the patch shortens the sequence by the same factor while keeping the signal.

Table~\ref{tab:sr_robust} compares the two at equal tokens per channel, row by row. FlexMLP enlarges the patch and keeps the recorded 100Hz signal; downsampling fixes the patch at 16 and lowers the rate. The last column is the gain of FlexMLP. Accuracy falls along both columns, since a shorter sequence carries less information either way. The gap between them tracks the bands in Table~\ref{tab:aasm_bands}. The two stay on par at 94 and 47 tokens. They separate at 10, where downsampling has removed the bands that carry the label, and a patch of 300 gives 72.21 against 66.17. Enlarging the patch therefore relieves the loss rather than removing it: accuracy still falls 10.82\% between 188 and 10 tokens, and what it avoids is giving up the bandwidth on top.

One set of embedding weights serves every patch size, resized to whichever one the heuristic picks. The FlexMLP column measures what that resizing costs. Every row sees the same recorded 100Hz signal, so moving from a patch of 16 to 32 changes only the resizing and halves the sequence, at a cost of 0.99\%. The resized kernel keeps the spectral structure that carries the label, even though its frequency response has changed. The larger drops further down the column come from the token budget itself.

\begin{table}[h]
\centering
\caption{\textbf{Enlarging the patch versus downsampling at equal sequence length.} (a) EEG rhythms retained at each sampling rate, on a 30s Sleep Stage window, with bands from the AASM manual~\citep{berry2020aasm}. (b) Sleep Stage linear probing accuracy at equal tokens per channel. Reference settings are marked in \colorbox{headerblue}{blue}.}
\label{tab:patch_analysis}

\begin{subtable}[t]{0.5\linewidth}
\centering
\caption{}
\label{tab:aasm_bands}
\resizebox{\linewidth}{!}{%
\begin{tabular}{lcccc}
\toprule
& \cellcolor{headerblue}\textbf{100Hz (native)} & \textbf{50Hz} & \textbf{25Hz} & \textbf{5Hz} \\
\midrule
Slow wave activity (0.5--2Hz) & \cellcolor{headerblue}kept & kept & kept & kept \\
Theta (4--8Hz) & \cellcolor{headerblue}kept & kept & kept & \textcolor{textred}{removed} \\
Alpha (8--13Hz) & \cellcolor{headerblue}kept & kept & partial & \textcolor{textred}{removed} \\
Sleep spindles (11--16Hz) & \cellcolor{headerblue}kept & kept & \textcolor{textred}{removed} & \textcolor{textred}{removed} \\
Beta (13--30Hz) & \cellcolor{headerblue}kept & partial & \textcolor{textred}{removed} & \textcolor{textred}{removed} \\
\bottomrule
\end{tabular}%
}
\end{subtable}
\hfill
\begin{subtable}[t]{0.47\linewidth}
\centering
\caption{}
\label{tab:sr_robust}
\resizebox{\linewidth}{!}{%
\begin{tabular}{lccccc}
\toprule
& \multicolumn{2}{c}{\textbf{FlexMLP}} & \multicolumn{2}{c}{\textbf{Downsampling}} & \\
\cmidrule(lr){2-3} \cmidrule(lr){4-5}
\textbf{Token budget} & \textbf{Patch size} & \textbf{Acc.} & \textbf{Rate} & \textbf{Acc.} & \textbf{Gain} \\
\midrule
\rowcolor{headerblue}
188 & 16 & 83.03 & 100Hz & 83.03 & -- \\
94 & 32 & 82.04 & 50Hz & 82.62 & \negval{-0.58} \\
47 & 64 & 79.01 & 25Hz & 78.96 & \posval{+0.05} \\
10 & 300 & \textbf{72.21} & 5Hz & 66.17 & \posval{+6.04} \\
\bottomrule
\end{tabular}%
}
\end{subtable}

\end{table}

\clearpage
\subsection{Uniformity and Alignment}
\label{sec:lunif}
\textbf{Setup.} To better understand the geometric properties of the embedding space, we follow \citet{wang2022unif}. The key idea is to view contrastive learning as balancing two geometric effects in the embedding space. Alignment measures whether a model pulls together embeddings of positive pairs that should represent the same underlying content, while uniformity measures whether embeddings of different samples remain spread out on the unit hypersphere instead of collapsing. In their original image setting, a positive pair is two augmented views of the same image, and negatives are views from other images in the batch. Good representations therefore require low alignment loss, meaning the two views of the same image land close, while also maintaining good uniformity, meaning the whole set of image embeddings does not crowd into a small region.

In sensor-language contrastive learning, the same definitions apply after replacing the notion of a positive pair. Here, the positive pair is a matched sensor segment \(x\) and its paired text description \(y\), and negatives are mismatched sensor-text pairs \((x_i, y_j)\) with \(i \neq j\). Alignment then becomes a direct measure of sensor-to-language coupling: if the representation is truly shared, embeddings of matched sensor and text should be close. Namely,
\begin{equation*}
    \mathcal{L}_{\text{align}} \triangleq \mathbb{E}_{(x,y)\sim \mathcal{D}^{+}}\!\left[\left\lVert f(x)-f(y)\right\rVert_{2}^{\alpha}\right],
    \quad \alpha>0.
\end{equation*}

\begin{figure}[h!]
\centering
\begin{minipage}[t]{0.56\linewidth}
    \centering
    \vspace{0pt}
    \includegraphics[width=\linewidth]{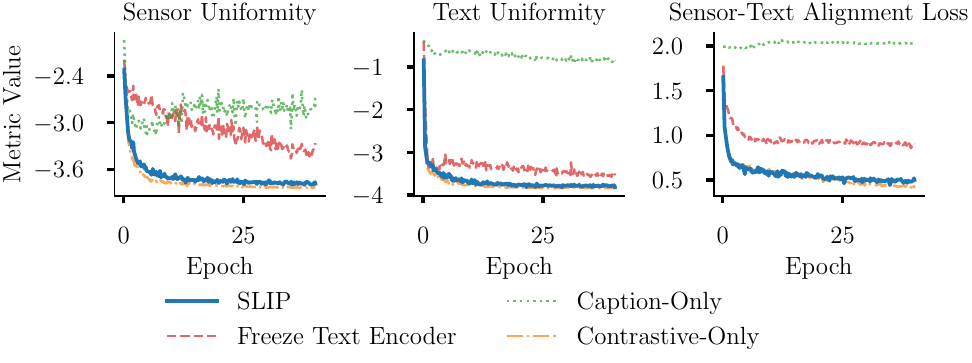}
    \captionof{figure}{\textbf{Sensor-Language representation geometry analysis.} \textbf{Sensor Uniformity (left)} and \textbf{Text Uniformity (middle)} quantify embedding dispersion on the unit hypersphere; \textbf{Sensor-Text Alignment (right)} measures mean distance between paired sensor and text embeddings. Lower is better.}
    \label{fig:contrastive_analysis}
\end{minipage}\hfill
\begin{minipage}[t]{0.40\linewidth}
    \vspace{0pt}
    \vspace*{-\intextsep}
    \begin{algorithm}[H]
    \caption{Implementation of alignment and uniformity.}
    \label{alg:align_unif}
    \lstset{style=mocov3}
    \begin{lstlisting}[language=python, escapechar=@, label=code:align_unif]
import torch
import torch.nn.functional as F

# bsz: batch size, d: latent dim
# x: (bsz, d) sensor emb
# y: (bsz, d) text emb

x = F.normalize(x, dim=1)
y = F.normalize(y, dim=1)

def lalign(x, y, alpha=2):
    return (x - y).norm(dim=1)\
        .pow(alpha).mean()

def lunif(x, t=2):
    sq = torch.pdist(x, p=2).pow(2)
    return sq.mul(-t).exp().mean().log()

align_loss   = lalign(x, y)
txt_unif     = lunif(y)
sensor_unif  = lunif(x)
    \end{lstlisting}
    \end{algorithm}
\end{minipage}
\vspace{-0.3cm}
\end{figure}

Uniformity becomes a check against collapse within each modality: sensor embeddings should remain spread across the sphere so different signals are distinguishable, and text embeddings should remain spread so different descriptions are distinguishable. The uniformity loss is defined as the logarithm of the average pairwise Gaussian potential:
\begin{align*}
    \mathcal{L}_{\text{unif}}(f; t)
    & \triangleq \log \mathbb{E}_{x, y \overset{\text{iid}}{\sim} \mathcal{D}} \left[ G_t(x, y) \right] \\
    & = \log \mathbb{E}_{x, y \overset{\text{iid}}{\sim} \mathcal{D}} \left[ e^{-t \|f(x) - f(y)\|_2^2} \right]
    , \quad t > 0.
\end{align*}
This is especially useful for sensor data because many windows can be similar and the model can otherwise reduce loss by relying on the stronger text branch, so tracking sensor uniformity alongside cross-modal alignment helps distinguish true cross-modal alignment from a degenerate solution. Altogether, we measure sensor-language alignment, sensor uniformity, and text uniformity with implementation in Alg.~\ref{alg:align_unif}.

\textbf{Result.} Figure~\ref{fig:contrastive_analysis} shows that the caption-only model suffers from poor sensor-text alignment and degraded sensor and text uniformity, consistent with representation collapse. Notably, when freezing the text encoder, sensor features spread more slowly, plateau earlier, and exhibit weaker sensor-text alignment. 

\clearpage
\subsection{Supplementary Results}
\vspace{-0.2cm}
\label{full_result}
\begin{table*}[h]
\caption{\textbf{Additional F1 scores.} Classification (LP) and retrieval (ZS), mirroring Table~\ref{tab:cls_ret}.}
\label{tab:f1_results}
\resizebox{\linewidth}{!}{%
\begin{tabular}{l|l|ccccccccccc|c}
\toprule
\textbf{Eval} & \textbf{Model} &
\textbf{WISDM} &
\textbf{UCIHAR} &
\textbf{Stroke} &
\textbf{Diabetes} &
\textbf{Hypertension} &
\textbf{Sleep Stage} &
\textbf{Heart Cond.} &
\textbf{WESAD} &
\textbf{StudentLife} &
\textbf{Obstacles} &
\textbf{Beijing AQI} &
\textbf{Average} \\
\midrule
\multirow{8}{*}{LP}
& Stat Feat
& 72.03 & 84.18 & 47.41 & 45.12 & 26.85 & 66.41 & 39.12 & 47.81 & \textbf{42.46} & 80.98 & 36.86 & 53.57 \\
& SimMTM
& 42.93 & 69.13 & 47.20 & 31.35 & \textbf{45.12} & 28.68 & 17.17 & 54.56 & 35.86 & 75.85 & 37.32 & 44.11 \\
& TFC
& 54.60 & 64.20 & 47.62 & 45.12 & 23.86 & 33.08 & 24.50 & 23.60 & 25.11 & 47.40 & 23.44 & 37.50 \\
& NormWear
& 70.60 & 76.88 & 52.30 & 44.90 & 32.99 & 72.25 & 47.52 & 63.02 & 32.24 & 82.82 & 48.58 & 56.74 \\
& Chronos2 
& 74.43 & 81.41 & 47.41 & 43.28 & 34.96 & \textbf{73.83} & 35.76 & 57.26 & 26.35 & 78.83 & 25.28 & 52.62 \\
& Chronos
& 80.06 & 81.73 & 47.35 & \textbf{57.70} & 35.54 & 66.07 & 48.89 & 51.72 & 28.17 & 74.61 & \textbf{61.17} & 57.47 \\
& Sundial Base
& 80.90 & 41.64 & 47.62 & 45.12 & 13.54 & 51.18 & 30.53 & 36.85 & 38.41 & 67.68 & 36.15 & 44.51 \\
& ChatTS & 75.36 & 57.31 & 47.62 & 45.12 & 13.93 & 69.70 & 42.36 & 57.31 & 36.19 & 76.12 & 29.85 & 50.08 \\
& SensorLM
& 76.17 & 84.03 & 50.15 & 52.06 & 37.12 & 70.90 & \textbf{50.06} & 69.14 & 42.31 & 82.88 & 52.24 & 60.64 \\
& \cellcolor{headerblue}\textbf{SLIP} &
\cellcolor{headerblue}\textbf{81.87} &
\cellcolor{headerblue}\textbf{84.58} &
\cellcolor{headerblue}\textbf{54.74}&
\cellcolor{headerblue}51.71 &
\cellcolor{headerblue}39.59 &
\cellcolor{headerblue}70.98 &
\cellcolor{headerblue}48.67 &
\cellcolor{headerblue}\textbf{71.84} &
\cellcolor{headerblue}40.62 &
\cellcolor{headerblue}\textbf{84.32} &
\cellcolor{headerblue}45.32 &
\cellcolor{headerblue}\underline{\textbf{61.17}} \\
\midrule \midrule
\multirow{2}{*}{ZS}
& NormWear
& 0.48 & 3.89 & \textbf{47.20} & 45.12 & 13.48 & 11.24 & 5.37 & 9.49 & 20.09 & 14.00 & 1.83 & 15.65 \\
& SensorLM
& 4.26 & 12.87 & 38.42 & 46.18 & 17.60 & 12.07 & \textbf{18.01} & 36.00 & 14.04 & 10.73 & 12.19 & 20.22 \\
& \cellcolor{headerblue}\textbf{SLIP}
& \cellcolor{headerblue}\textbf{5.47}
& \cellcolor{headerblue}\textbf{13.80}
& \cellcolor{headerblue}46.34
& \cellcolor{headerblue}\textbf{45.19}
& \cellcolor{headerblue}\textbf{17.15}
& \cellcolor{headerblue}\textbf{29.22}
& \cellcolor{headerblue}5.36
& \cellcolor{headerblue}\textbf{46.02}
& \cellcolor{headerblue}\textbf{24.97}
& \cellcolor{headerblue}\textbf{15.33}
& \cellcolor{headerblue}\textbf{13.81}
& \cellcolor{headerblue}\underline{\textbf{23.88}} \\
\bottomrule
\end{tabular}%
}
\end{table*}
\begin{table*}[h]
\caption{\textbf{Additional supplementary results.} Classification (LP) and retrieval (ZS), mean $\pm$ std over 5 random seeds.}
\label{tab:kfold_res}
\resizebox{\linewidth}{!}{%
\begin{tabular}{l|l|ccccccccccc}
\toprule
\textbf{Eval} & \textbf{Model} &
\textbf{WISDM} &
\textbf{UCIHAR} &
\textbf{Stroke} &
\textbf{Diabetes} &
\textbf{Hypertension} &
\textbf{Sleep Stage} &
\textbf{Heart Cond.} &
\textbf{WESAD} &
\textbf{StudentLife} &
\textbf{Obstacles} &
\textbf{Beijing AQI} \\
\midrule
\multirow{11}{*}{LP}
& Stat Feat
& 77.14 $\pm$ 0.25
& 85.62 $\pm$ 0.38
& 90.15 $\pm$ 0.00
& 82.22 $\pm$ 0.00
& 37.73 $\pm$ 1.30
& 77.33 $\pm$ 0.02
& 60.99 $\pm$ 0.33
& 68.52 $\pm$ 0.77
& 48.26 $\pm$ 1.37
& 81.69 $\pm$ 0.80
& 68.81 $\pm$ 0.51 \\
& SimMTM
& 33.76 $\pm$ 0.08
& 70.52 $\pm$ 0.24
& 89.39 $\pm$ 0.00
& 82.22 $\pm$ 0.00
& 37.88 $\pm$ 1.07
& 41.76 $\pm$ 0.07
& 55.90 $\pm$ 0.08
& 62.51 $\pm$ 0.46
& 50.64 $\pm$ 0.69
& 36.37 $\pm$ 0.98
& 71.26 $\pm$ 0.26 \\
& TFC
& 44.11 $\pm$ 0.11
& 59.95 $\pm$ 0.13
& 89.85 $\pm$ 0.61
& 80.74 $\pm$ 0.47
& 40.00 $\pm$ 0.57
& 44.88 $\pm$ 0.05
& 58.11 $\pm$ 0.25
& 55.43 $\pm$ 0.46
& 43.67 $\pm$ 1.70
& 33.66 $\pm$ 0.62
& 45.60 $\pm$ 0.46 \\
& NormWear
& 70.82 $\pm$ 0.03
& 79.37 $\pm$ 0.32
& 90.45 $\pm$ 0.61
& 82.37 $\pm$ 0.30
& 40.45 $\pm$ 1.13
& 82.92 $\pm$ 0.18
& 69.53 $\pm$ 0.07
& 80.27 $\pm$ 0.49
& 49.17 $\pm$ 1.24
& 83.38 $\pm$ 0.32
& 74.40 $\pm$ 0.38 \\
& Chronos2
& 75.18 $\pm$ 0.07
& 75.38 $\pm$ 0.51
& 87.12 $\pm$ 0.96
& 73.33 $\pm$ 3.18
& 38.79 $\pm$ 0.88
& \textbf{83.74} $\pm$ 0.14
& 61.20 $\pm$ 0.19
& 67.26 $\pm$ 0.75
& 47.89 $\pm$ 0.69
& 84.14 $\pm$ 0.23
& 51.74 $\pm$ 0.63 \\
& Chronos
& 81.19 $\pm$ 0.13
& 84.04 $\pm$ 0.30
& 88.48 $\pm$ 2.00
& 80.59 $\pm$ 1.36
& 45.30 $\pm$ 0.88
& 76.69 $\pm$ 0.14
& \textbf{73.27} $\pm$ 0.18
& 66.46 $\pm$ 0.18
& 49.54 $\pm$ 1.00
& 75.45 $\pm$ 0.28
& \textbf{80.48} $\pm$ 0.14 \\
& Sundial Base
& 41.39 $\pm$ 0.12
& 55.84 $\pm$ 0.29
& 77.27 $\pm$ 0.83
& 72.00 $\pm$ 5.03
& 37.73 $\pm$ 3.16
& 70.19 $\pm$ 0.14
& 61.26 $\pm$ 0.12
& 59.64 $\pm$ 0.40
& 47.53 $\pm$ 2.20
& 68.04 $\pm$ 0.23
& 67.85 $\pm$ 0.26 \\
& MOMENT
& 69.99 $\pm$ 0.08
& 72.71 $\pm$ 0.41
& 82.88 $\pm$ 6.58
& 78.22 $\pm$ 3.30
& 45.91 $\pm$ 1.03
& 80.06 $\pm$ 0.03
& 72.85 $\pm$ 0.07
& 71.75 $\pm$ 0.28
& 48.26 $\pm$ 1.10
& \textbf{85.68} $\pm$ 0.16
& 61.84 $\pm$ 0.79 \\
& ChatTS
& 75.66 $\pm$ 0.00
& 56.24 $\pm$ 0.00
& 90.91 $\pm$ 0.00
& 82.22 $\pm$ 0.00
& 44.70 $\pm$ 0.00
& 80.65 $\pm$ 0.08
& 67.93 $\pm$ 0.21
& 77.04 $\pm$ 0.18
& 50.28 $\pm$ 0.90
& 77.75 $\pm$ 0.65
& 61.23 $\pm$ 1.84 \\
& SensorLM
& 76.57 $\pm$ 0.18
& \textbf{88.83} $\pm$ 0.29
& 85.45 $\pm$ 1.11
& 75.11 $\pm$ 0.76
& 44.24 $\pm$ 0.61
& 80.74 $\pm$ 0.05
& 71.05 $\pm$ 0.08
& 79.55 $\pm$ 0.46
& 48.61 $\pm$ 0.90
& 83.22 $\pm$ 0.31
& 76.04 $\pm$ 0.33 \\
& \cellcolor{headerblue}\textbf{SLIP}
& \cellcolor{headerblue}\textbf{82.28} $\pm$ 0.10
& \cellcolor{headerblue}85.67 $\pm$ 0.26
& \cellcolor{headerblue}\textbf{91.36} $\pm$ 0.67
& \cellcolor{headerblue}\textbf{83.11} $\pm$ 0.55
& \cellcolor{headerblue}\textbf{47.58} $\pm$ 0.57
& \cellcolor{headerblue}82.62 $\pm$ 0.09
& \cellcolor{headerblue}71.59 $\pm$ 0.19
& \cellcolor{headerblue}\textbf{81.79} $\pm$ 0.61
& \cellcolor{headerblue}\textbf{54.31} $\pm$ 1.58
& \cellcolor{headerblue}85.32 $\pm$ 0.35
& \cellcolor{headerblue}76.66 $\pm$ 0.27 \\
\midrule \midrule
\multirow{3}{*}{ZS}
& NormWear
& 3.91 $\pm$ 0.66
& 12.98 $\pm$ 2.15
& 89.40 $\pm$ 3.74
& \textbf{82.22 }$\pm$ 7.18
& \textbf{36.38} $\pm$ 8.38
& \textbf{39.09} $\pm$ 0.73
& 15.52 $\pm$ 1.26
& 16.58 $\pm$ 6.43
& \textbf{43.12} $\pm$ 13.02
& 24.30 $\pm$ 2.03
& 2.40 $\pm$ 2.34 \\
& SensorLM
& \textbf{9.66} $\pm$ 0.26
& 26.05 $\pm$ 0.54
& 45.00 $\pm$ 3.57
& 67.11 $\pm$ 2.46
& 28.64 $\pm$ 2.64
& 15.96 $\pm$ 0.25
& \textbf{48.64} $\pm$ 0.32
& 38.12 $\pm$ 2.02
& 19.08 $\pm$ 1.35
& 25.53 $\pm$ 0.38
& 12.90 $\pm$ 1.64 \\
& \cellcolor{headerblue}\textbf{SLIP}
& \cellcolor{headerblue}7.45 $\pm$ 0.14
& \cellcolor{headerblue}\textbf{27.26} $\pm$ 0.46
& \cellcolor{headerblue}\textbf{90.91} $\pm$ 0.00
& \cellcolor{headerblue}75.56 $\pm$ 2.65
& \cellcolor{headerblue}35.76 $\pm$ 0.30
& \cellcolor{headerblue}34.58 $\pm$ 0.17
& \cellcolor{headerblue}16.08 $\pm$ 0.19
& \cellcolor{headerblue}\textbf{40.45} $\pm$ 0.87
& \cellcolor{headerblue}42.02 $\pm$ 0.69
& \cellcolor{headerblue}\textbf{30.08} $\pm$ 0.13
& \cellcolor{headerblue}\textbf{22.25} $\pm$ 1.67 \\
\bottomrule
\end{tabular}%
}
\end{table*}
\begin{table*}[h]
\caption{\textbf{Ablation Studies Linear Probing Performance.} Reported Top-1 accuracy on 11 datasets.}
\label{tab:lp_res}
\resizebox{\linewidth}{!}{%
\begin{tabular}{l|ccccccccccc|c}
\toprule
\textbf{Model} &
\textbf{WISDM} &
\textbf{UCIHAR} &
\textbf{Stroke} &
\textbf{Diabetes} &
\textbf{Hypertension} &
\textbf{Sleep Stage} &
\textbf{Heart Cond.} &
\textbf{WESAD} &
\textbf{StudentLife} &
\textbf{Obstacles} &
\textbf{Beijing AQI} &
\textbf{All Avg} \\
\midrule
\rowcolor{headerblue}
\textbf{\systemname{}} &
\textbf{82.36} & 86.25 & 91.67 & 84.44 & \textbf{50.00} & 82.04 & 70.79 & 82.96 & 54.13 & \textbf{86.45} & 77.47 & \textbf{77.14} \\
\midrule \toprule
Contrastive-Only &
79.14 & 85.37 & 90.91 & 84.44 & 40.91 & 81.10 & 68.12 & 82.96 & 50.46 & 84.65 & 74.40 & 74.77 \\
Caption-Only &
82.11 & 84.36 & 90.91 & 83.70 & 37.12 & 81.96 & 69.70 & 76.68 & 46.79 & 85.42 & 78.50 & 74.30 \\
Random Paired &
38.16 & 65.45 & 90.91 & 82.96 & 41.67 & 59.62 & 55.39 & 52.47 & 52.47 & 74.93 & 69.62 & 62.15\\ 
\midrule \toprule
w/o \textit{FlexMLP} ($ps=16$) &
76.80 & \textbf{89.16} & 87.88 & 82.22 & 47.73 & 82.55 & 70.73 & 72.20 & 46.79 & 84.91 & 74.74 & 74.16 \\
\midrule \toprule
w/ Group Attention &
79.45 & 84.36 & \textbf{92.42} & \textbf{85.19} & 48.48 & 81.67 & \textbf{72.12} & 83.41 & \textbf{56.88} & 85.94 & 77.47 & 77.04 \\
\midrule \toprule
Freeze Text Encoder &
76.75 & 85.12 & 83.33 & 79.26 & 40.91 & 79.21 & 66.55 & \textbf{87.44} & 54.13 & 82.35 & 74.06 & 73.56\\
\midrule \toprule
Llama3.2-1B Backbone &
82.16 & 86.63 & 88.79 & 81.63 & 38.03 & \textbf{82.75} & 71.02 & 80.36 & 51.19 & 85.52 & 77.27 & 75.03 \\
\midrule \toprule
SLIP\textsubscript{Small} (40M) &
80.54 & 81.84 & 90.91 & 83.70 & 42.42 & 80.74 & 68.67 & 81.17 & 49.54 & 84.91 & \textbf{78.84} & 74.84 \\
\bottomrule
\end{tabular}%
}
\end{table*}

\begin{table*}[h]
\caption{\textbf{Ablation Studies Retrieval Performance.} Top-1 accuracy on 11 datasets.}
\label{tab:retrieval_res}
\resizebox{\linewidth}{!}{%
\begin{tabular}{l|ccccccccccc|c}
\toprule
\textbf{Model} &
\textbf{WISDM} &
\textbf{UCIHAR} &
\textbf{Stroke} &
\textbf{Diabetes} &
\textbf{Hypertension} &
\textbf{Sleep Stage} &
\textbf{Heart Cond.} &
\textbf{WESAD} &
\textbf{StudentLife} &
\textbf{Obstacles} &
\textbf{Beijing AQI} &
\textbf{All Avg} \\
\midrule
\rowcolor{headerblue}
\textbf{\systemname{}} &
7.45 & \textbf{27.62} & 86.40 & 74.07 & 34.85 & 34.83 & 15.39 & \textbf{56.95} & 41.28 & 29.92 & 24.23 & \textbf{39.36} \\
\midrule \toprule
Contrastive-Only &
9.50 & 19.67 & 79.55 & 80.74 & 24.24 & 39.05 & 10.97 & 52.47 & 40.37 & 30.18 & 19.11 & 36.90 \\
Caption-Only  &
5.82 & 6.56 & 78.03 & 17.78 & \textbf{38.64} & 20.87 & 7.21 & 21.52 & 42.20 & 29.16 & 8.53 & 25.12 \\
Random Paired &
5.09 & 4.04 & 81.06 & 17.78 & 9.09 & 22.20 & 3.39 & 16.59 & 24.77 & 18.16 & 44.03 & 22.38\\
\midrule \toprule
w/ Group Attention &
8.30 & 27.11 & 84.09 & \textbf{82.22} & 36.36 & 26.28 & \textbf{26.85} & 52.02 & \textbf{44.95} & 24.04 & 19.11 & 39.21 \\
\midrule \toprule
w/o \textit{FlexMLP} ($ps=16$) &
5.86 & 13.75 & \textbf{90.15} & 59.26 & 29.54 & 29.02 & 8.79 & 40.81 & 30.28 & 30.18 & 45.05 & 34.79 \\
\midrule \toprule
Freeze Text Encoder &
\textbf{12.02} & 17.28 & 70.46 & \textbf{82.22} & 34.85 & \textbf{39.48} & 13.15 & 23.32 & 36.70 & \textbf{42.20} & 20.82 & 35.68 \\
\midrule \toprule
Llama3.2-1B Backbone &
8.81 & 24.21 & 63.18 & \textbf{82.22} & 36.36 & 32.51 & 13.01 & 30.94 & 42.57 & 31.30 & 36.72 & 36.53 \\
\midrule \toprule
SLIP\textsubscript{Small} (40M) &
11.27 & 9.84 & 43.18 & 32.59 & 36.36 & 33.01 & 14.67 & 22.42 & 20.18 & 20.46 & \textbf{49.83} & 26.71 \\
\bottomrule
\end{tabular}%
}
\end{table*}

\begin{table*}[h]
\centering
\caption{\textbf{Ablation Studies SFT performance.}  Top-1 accuracy on four sensor QA benchmarks.}
\label{tab:sft_ablation}
\resizebox{0.6\linewidth}{!}{%
\begin{tabular}{l|ccccc}
\toprule
\textbf{Model} &
\textbf{TSQA} &
\textbf{HAR-CoT} &
\textbf{Sleep-CoT} &
\textbf{ECG-QA-CoT} &
\textbf{Avg} \\
\midrule
\rowcolor{headerblue}
\textbf{\systemname{}} &
\textbf{99.75} & \textbf{64.35} & 74.19 & 37.18 & 68.87 \\
\midrule \toprule
Caption-Only &
99.04 & 49.16 & 69.14 & 34.07 & 62.85 \\
Contrastive-Only &
95.88 & 48.57 & 29.57 & 38.21 & 53.06 \\
Random Paired &
74.40 & 27.93 & 21.51 & 33.16 & 39.25 \\
\midrule \toprule
w/ Group Attention &
99.27 & 60.41 & 69.14 & 37.05 & 66.47 \\
\midrule \toprule
w/o \textit{FlexMLP} ($ps=16$)  &
98.65 & 61.66 & 69.03 & 38.17 & 66.88 \\
\midrule \toprule
Freeze Text Encoder &
99.38 & 47.00 & 61.18 & 34.62 & 60.55 \\
\midrule \toprule
Llama3.2-1B Backbone &
99.60 & 62.65 & \textbf{74.25} & \textbf{43.47} & \textbf{69.99} \\
\midrule \toprule
SLIP\textsubscript{Small} (40M) &
99.35 & 50.58 & 64.09 & 25.70 & 59.93 \\
\bottomrule
\end{tabular}%
}
\end{table*}


\begin{table*}[h]
\centering
\caption{\textbf{Sensor captioning and question answering results with larger-LLM baselines.} Extends Table~\ref{tab:sft_res} with ITFormer-0.5B and ChatTS-8B, which use larger LLMs than the Gemma-3-270M backbone shared by OpenTSLM, SensorLM, and \systemname{}; they are shown in \textcolor{mygray}{gray} and excluded from highlighting. Metrics and protocol follow Table~\ref{tab:sft_res}. Best results are in \textbf{bold} and second best are \underline{underlined}.}
\label{tab:sft_res_full}
\resizebox{\linewidth}{!}{%
\begin{tabular}{l|ccccc|cccc}
\toprule
\multirow{2}{*}{\textbf{Model}} &
\multicolumn{5}{c|}{\textbf{Sensor Captioning (M4)}} &
\textbf{TSQA} & \textbf{HAR-CoT} & \textbf{Sleep-CoT} & \textbf{ECG-QA-CoT} \\
\cmidrule(lr){2-6} \cmidrule(lr){7-10}
 &
\textbf{BLEU@4} & \textbf{METEOR} & \textbf{ROUGE-L} & \textbf{SBERTSim} & \textbf{BERTScore} &
\textbf{Accuracy} & \textbf{Accuracy} & \textbf{Accuracy} & \textbf{Accuracy} \\
\midrule
\color{mygray}ITFormer & \color{mygray}0.1641 & \color{mygray}0.3708 & \color{mygray}0.3182 & \color{mygray}0.8748 & \color{mygray}0.9001 &
\color{mygray}89.35 & \color{mygray}41.29 & \color{mygray}43.99 & \color{mygray}38.01 \\
\color{mygray}ChatTS & \color{mygray}0.1252 & \color{mygray}0.3518 & \color{mygray}0.2950 & \color{mygray}0.8500 & \color{mygray}0.8906 &
\color{mygray}96.25 & \color{mygray}41.16 & \color{mygray}55.22 & \color{mygray}34.81 \\
\midrule
OpenTSLM-SP &
0.0026 & 0.0456 & 0.0255 & 0.1551 & 0.7250 &
26.79 & 0.55 & 5.91 & 1.11 \\
OpenTSLM-Flamingo &
0.1141 & 0.3210 & 0.2894 & 0.7990 & 0.8858 &
78.12 & \underline{63.43} & 68.49 & \underline{35.50} \\
SensorLM &
\underline{0.2008} & \textbf{0.4151} & \underline{0.3595} & \textbf{0.9014} & \underline{0.9024} &
\underline{95.73} & 60.80 & \underline{73.12} & 30.29 \\
\rowcolor{headerblue}
\textbf{\systemname{}} &
\textbf{0.2046} & \underline{0.4027} & \textbf{0.3658} & \underline{0.8977} & \textbf{0.9077} &
\textbf{99.75} & \textbf{64.35} & \textbf{74.19} & \textbf{37.18} \\
\bottomrule
\end{tabular}%
}
\end{table*}

\FloatBarrier
\clearpage
\subsection{Additional Examples of Question Answering}
\label{add_res}
\begin{figure}[h]
    \centering
    \includegraphics[width=0.7\linewidth]{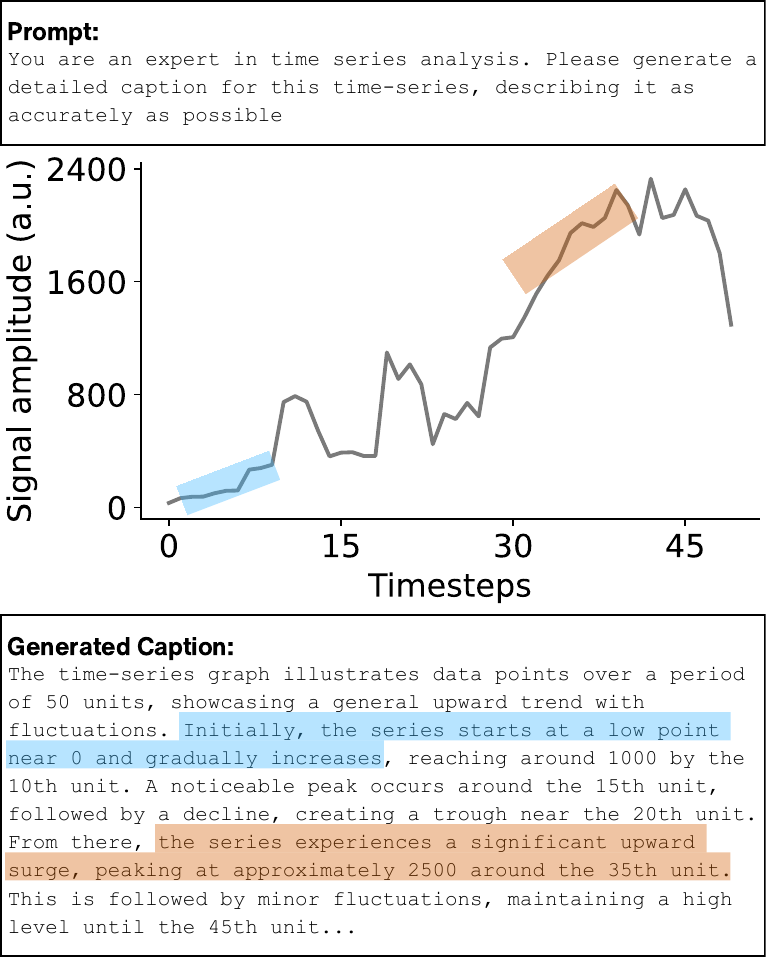}
    \caption{\textbf{A qualitative example of sensor captioning from the M4 dataset.} Highlighted segments denote salient local trends referenced in the generated caption.}
    \label{fig:m4qa_demo}
\end{figure}
\begin{figure}
    \centering
    \includegraphics[width=0.5\linewidth]{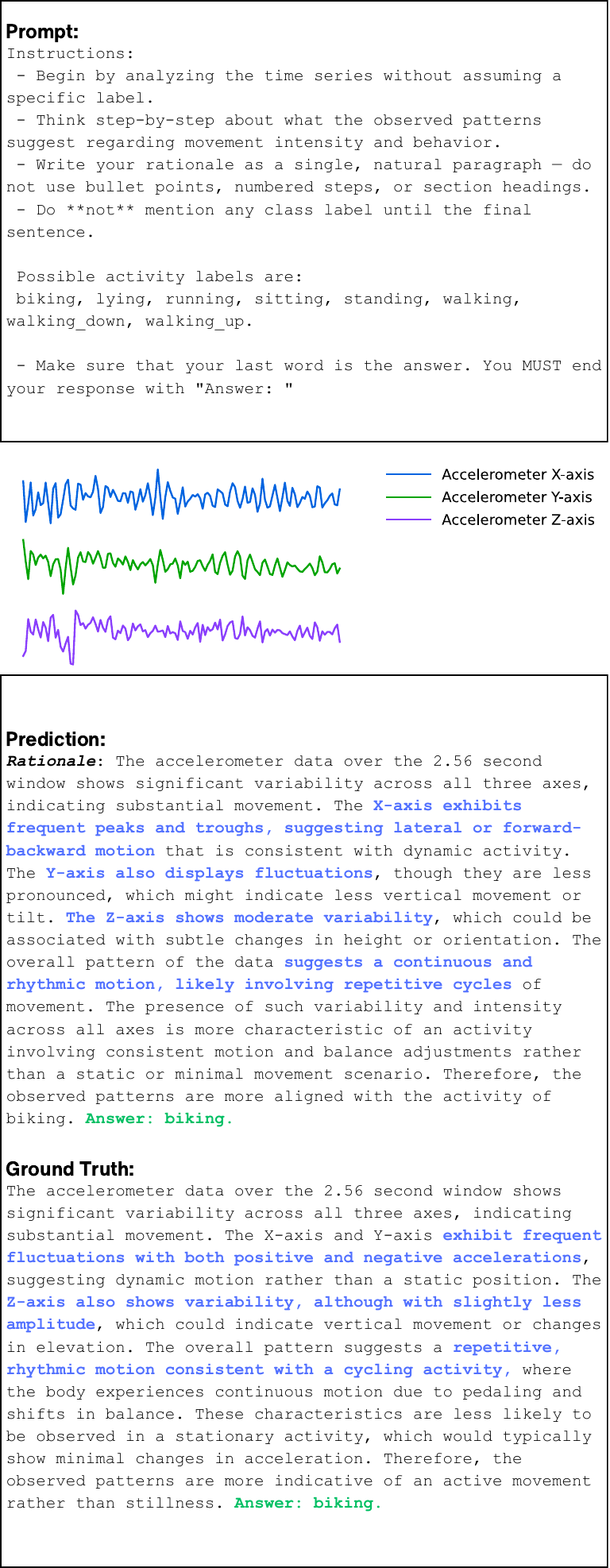}
    \caption{A qualitative example of sensor question answering from the HAR-CoT dataset.}
    \label{fig:har_cot_demo}
\end{figure}
\begin{figure}
    \centering
    \includegraphics[width=0.5\linewidth]{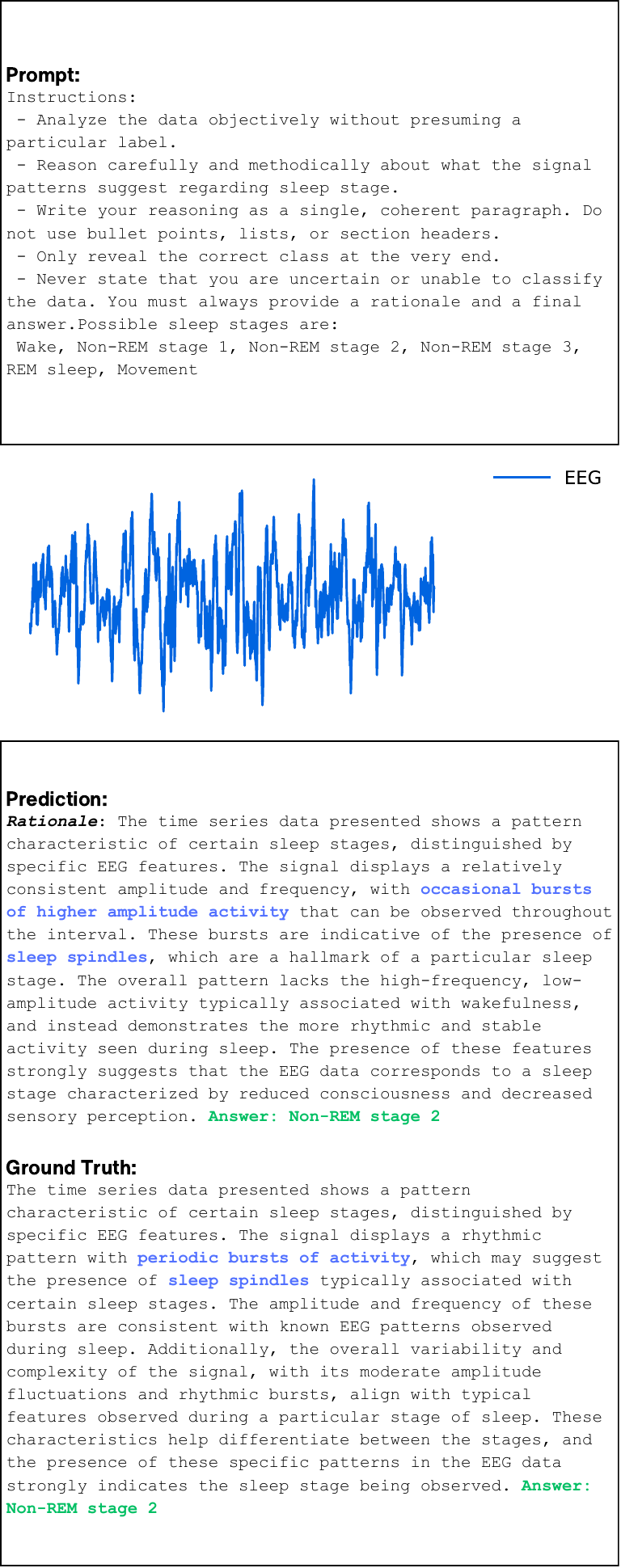}
    \caption{A qualitative example of sensor question answering from the Sleep-CoT dataset.}
    \label{fig:sleep_cot_demo}
\end{figure}
\begin{figure}
    \centering
    \includegraphics[width=0.55\linewidth]{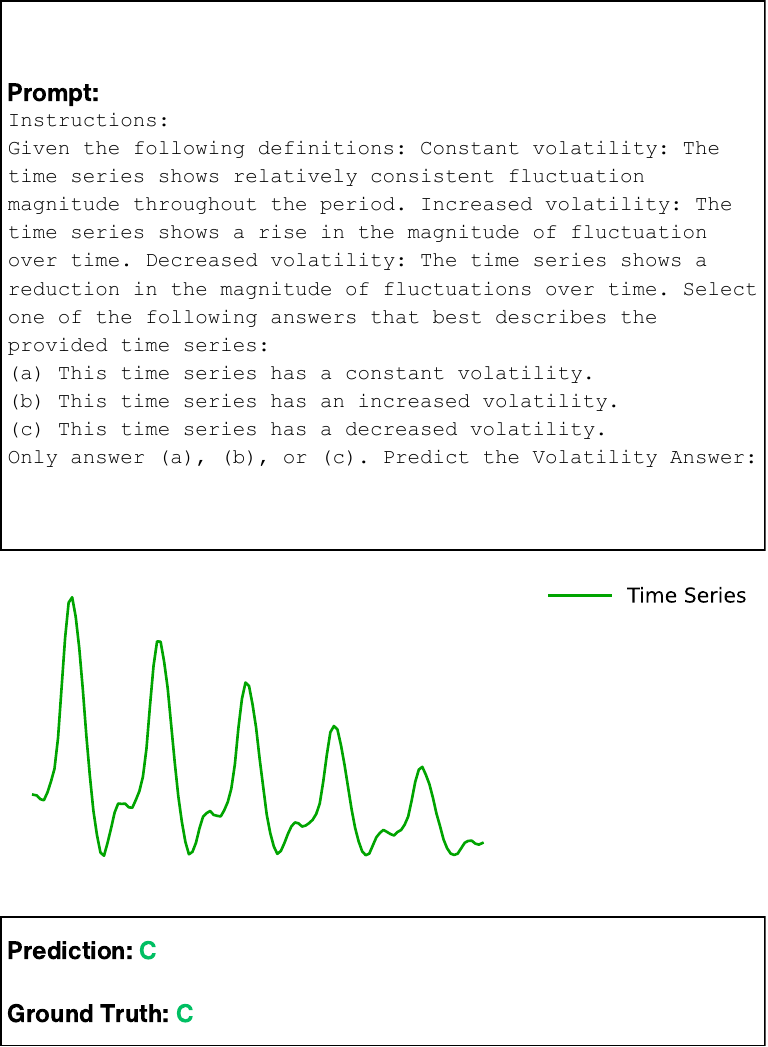}
    \caption{A qualitative example of sensor question answering from the TSQA dataset.}
    \label{fig:tsqa_demo}
\end{figure}
\begin{figure}
    \centering
    \includegraphics[width=0.35\linewidth]{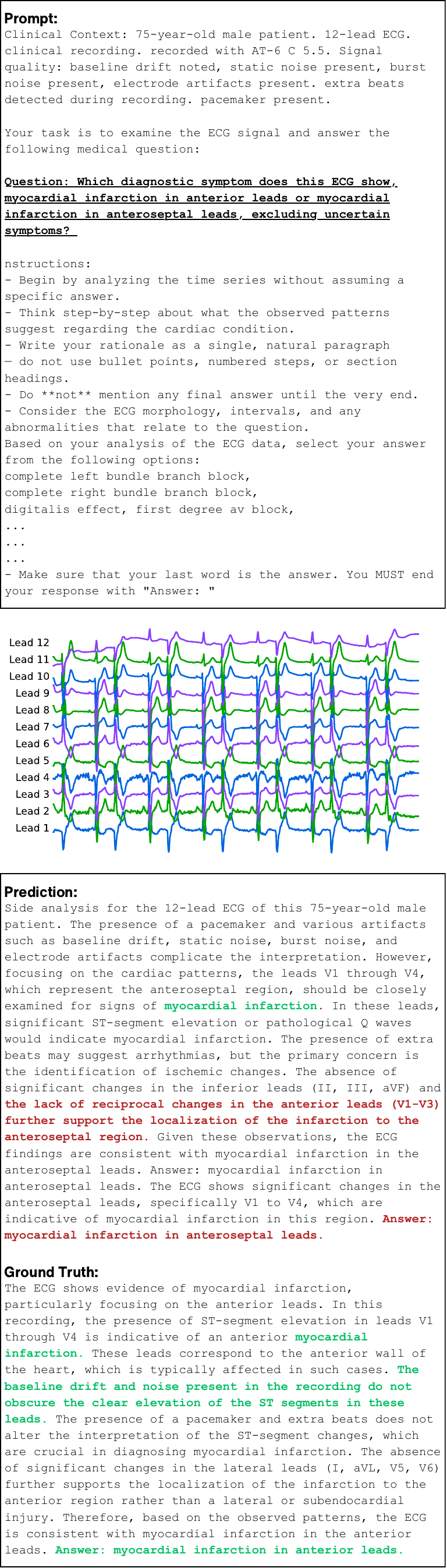}
    \caption{A qualitative failure example of sensor question answering from the ECG-QA-CoT dataset.}
    \label{fig:ecg_failure}
\end{figure}


\end{document}